%% file: iclr2023_conference.tex
\documentclass{article} 
\usepackage{iclr2023_conference,times}
\usepackage{booktabs}
\usepackage{multirow}
\usepackage{soul}
\input{math_commands.tex}

\usepackage{pgfplots}
\usepackage{hyperref}
\usepackage{url}
\usepackage[most]{tcolorbox}
\usepackage{multicol}

\usepackage{physics}
\usepackage{verbatim}
\usepackage{amssymb}
\usepackage{pifont}
\usepackage{ulem}
\usepackage{mathrsfs}

\PassOptionsToPackage{prologue,dvipsnames}{xcolor}

\newcommand{\cmark}{\ding{51}}%
\newcommand{\xmark}{\ding{55}}%

\definecolor{codegreen}{rgb}{0,0.6,0}
\definecolor{codegray}{rgb}{0.5,0.5,0.5}
\definecolor{codepurple}{rgb}{0.58,0,0.82}
\definecolor{backcolour}{rgb}{0.95,0.95,0.92}
\usepackage{listings}

\lstdefinestyle{mystyle}{
  commentstyle=\color{codegreen},
  keywordstyle=\color{magenta},
  numberstyle=\tiny\color{codegray},
  stringstyle=\color{codepurple},
  basicstyle=\ttfamily\footnotesize,
  breakatwhitespace=false,         
  breaklines=true,                 
  captionpos=b,                    
  keepspaces=false,                                 
  showspaces=false,                
  showstringspaces=false,
  showtabs=false,                  
  tabsize=2
}

\newcommand\blfootnote[1]{%
  \begingroup
  \renewcommand\thefootnote{*}\footnote{#1}%
  \addtocounter{footnote}{-1}%
  \endgroup
}

\newcommand{\gptcode}{GPT4-Code}
\title{Solving Challenging Math Word Problems Using GPT-4 Code Interpreter with Code-based Self-Verification}

\author{\centerline{Aojun Zhou\textsuperscript{1}\textsuperscript{\thanks{Equal contribution.}}
\quad Ke Wang\textsuperscript{2}\textsuperscript{\blfootnote{Equal contribution.}} 
\quad Zimu Lu\textsuperscript{3}\textsuperscript{\blfootnote{Equal contribution.}} 
\quad Weikang Shi\textsuperscript{4}\textsuperscript{\blfootnote{Equal contribution.}} 
\quad Sichun Luo\textsuperscript{5}\textsuperscript{\blfootnote{Equal contribution.}}  
\quad Zipeng Qin\textsuperscript{1}} \\ 
\centerline{\textbf{Shaoqing Lu \textsuperscript{6}} 
\quad \textbf{Anya Jia \textsuperscript{7}} 
\quad \textbf{Linqi Song\textsuperscript{5}} 
\quad \textbf{Mingjie Zhan\textsuperscript{1}}\textsuperscript{\normalsize \thanks{Project lead.}} 
\quad \textbf{Hongsheng Li\textsuperscript{1}}\textsuperscript{\normalsize \thanks{Corresponding author.}}} \\\\
\centerline{\textsuperscript{1}Multimedia Laboratory (MMLab), The Chinese University of Hong Kong} \\
\centerline{\textsuperscript{2}Nanjing University \quad  \textsuperscript{3}University of Science and Technology of China} \\
\centerline{\textsuperscript{4}Tsinghua University  \quad \textsuperscript{5}City University of Hong Kong} \\ 
\centerline{\textsuperscript{6}Changsha University of Science and Technology\quad\textsuperscript{7}Tufts University} \\
\centerline{\texttt{\{aojunzhou, wangk.gm, sichunluo2, zmjdll\}@gmail.com}} \\ 
\centerline{\texttt{luzimu@mail.ustc.edu.cn} \quad \texttt{shiwk20@mails.tsinghua.edu.cn}}\\
\centerline{\texttt{linqi.song@cityu.edu.hk} \quad \texttt{hsli@ee.cuhk.edu.hk}}
}

\iclrfinalcopy
\begin{document}

\maketitle

\begin{abstract}

Recent progress in large language models (LLMs) like GPT-4 and PaLM-2 has brought significant advancements in addressing math reasoning problems. In particular, OpenAI's latest version of GPT-4, known as GPT-4 Code Interpreter, shows remarkable performance on challenging math datasets. In this paper, we explore the effect of code on enhancing LLMs' reasoning capability by introducing different constraints on the \textit{Code Usage Frequency} of GPT-4 Code Interpreter. We found that its success can be largely attributed to its powerful skills in generating and executing code, evaluating the output of code execution, and rectifying its solution when receiving unreasonable outputs. Based on this insight, we propose a novel and effective prompting method, explicit \uline{c}ode-based \uline{s}elf-\uline{v}erification~(CSV), to further boost the mathematical reasoning potential of GPT-4 Code Interpreter. This method employs a zero-shot prompt on GPT-4 Code Interpreter to encourage it to use code to self-verify its answers. In instances where the verification state registers as ``False'', the model shall automatically amend its solution, analogous to our approach of rectifying errors during a mathematics examination. Furthermore, we recognize that the states of the verification result indicate the confidence of a solution, which can improve the effectiveness of majority voting. With GPT-4 Code Interpreter and CSV, we achieve an impressive zero-shot accuracy on MATH dataset \textbf{(53.9\% $\to$ 84.3\%)}. 

\end{abstract}
\section{Introduction}

Large language models (LLMs)~\citep{brown2020language, openai2023gpt4, anil2023palm} have shown impressive success in various tasks, such as common sense understanding and code generation. However, they still fall short in mathematical reasoning, often producing nonsensical or inaccurate content and struggling with complex calculations. Previous attempts to tackle these challenges include the Chain-of-Thought (CoT)~\citep{wei2022chain} framework,  which enhances LLMs' logical reasoning abilities by generating intermediate steps in their reasoning process. Additionally, PAL~\citep{gao2023pal} introduces a novel approach by using the Python programming interpreter to improve computational accuracy.

In recent advancements, OpenAI has unveiled an improved version of GPT-4, namely the GPT-4 Code Interpreter\footnote{\url{https://openai.com/blog/chatgpt-plugins##code-interpreter}}\footnote{\url{https://chat.openai.com/?model=\gptcode-interpreter}} or \textit{\gptcode}, which is proficient at providing logical natural language reasoning, alongside step-by-step Python code. Notably, it can generate and execute code incrementally, and subsequently present the executed code's output back to the LLM. The addition of code generation and execution to natural language outputs has shown promising results in solving mathematical reasoning problems. Our initial experiments show that \gptcode~achieved an impressive
 zero-shot accuracy of 69.7\% on the challenging MATH dataset~\citep{hendrycks2021measuring}, marking a significant improvement of 27.5\% over GPT-4's performance (42.2\%).

While \gptcode~has demonstrated proficiency in solving math problems, there has been a notable absence of systematic analysis focusing on understanding and further enhancing its mathematical problem-solving abilities. A key distinction between \gptcode~and its predecessor, GPT-4, lies in \gptcode 's ability to \textit{automatically generate and execute code}. Therefore,  this paper presents pilot experiments that investigate \gptcode 's code generation and execution mechanism using specific code-constrained prompts. The analysis reveals that \gptcode 's strong performance is not solely due to its code generation and execution abilities, but also its capacity to adjust its problem-solving strategies based on feedback from code execution—a process we term \textbf{self-debugging} (illustrated in Tab.~\ref{tab:baseline_example1} and Tab.~\ref{tab:baseline_example2}). Due to the fact that code generation evolves its reasoning step-by-step and performs self-debugging after code execution errors, there is an increased frequency of code usage. Hence, we introduce the concept of Code Usage Frequency to differentiate these unique prompting strategies to quantitatively analyze the impact of code-constrained prompts on \gptcode~for mathematical problem-solving. 

The step-by-step code generation and self-debugging mechanisms highlight the critical role of code in mathematical problem-solving. Nevertheless, the self-debugging mechanism only verifies each step of the generated code while lacks the verification of the reasoning steps and the final answer, which has been demonstrated to be of vital importance to the math problem-solving abilities of LLMs~\citep{cobbe2021gsm8k, lightman2023lets, weng2023large}. 

We therefore ask the question: \textit{can we fully exploit the code generation and self-debugging mechanisms in GPT4-code, so that it can automatically verify and correct its solutions, without extra assistance from other models or users?}

To answer this question, we propose a simple yet effective prompting technique termed the \textbf{explicit \uline{c}ode-based \uline{s}elf-\uline{v}erification} (CSV), which guides \gptcode~to generate additional code that verifies the answer and adjusts the reasoning steps if there's a flaw in reasoning. Unlike previous methods that rely on external language models for verification~\citep{lightman2023lets,cobbe2021gsm8k}, our approach leverages \gptcode's inherent strengths. This approach offers two key benefits: (1) When the verification indicates an answer is \textbf{False}, \gptcode~can rectify its prior solution and provide an improved alternative. (2) Solutions verified as \textbf{True} tend to be more reliable, akin to human problem-solving. However, even if a solution is self-verified as \textbf{False}, we do not directly abandon it. Instead, we propose a weighted majority voting strategy that incorporates the code-based solution verification results, as opposed to relying exclusively on the frequency of answers. We assign different weights to the solutions according to their verification states, reflecting the solutions' varying levels of reliability. In alignment with the Code Usage Frequency analysis from our pilot experiments, our explicit code-based self-verification prompt boosts \gptcode's accuracy in mathematical problem-solving with increased code usage.

Empirical study demonstrates the effectiveness of our proposed framework on the MATH, GSM8K, and MMLU-Math datasets using \gptcode. Our approach achieves an impressive accuracy of \textbf{84.32\%} on the MATH dataset, greatly outperforming the base \gptcode~and previous state-of-the-art methods. Additionally, we are making our experimental data on the MMLU-Math and MATH datasets publicly available, enabling result replication and facilitating fine-tuning of open-source LLM models (e.g., LLaMA 2~\citep{touvron2023llama}) to further enhance mathematical problem-solving capabilities with the assistance of code generation.

This paper's main contributions can be summarized in three key aspects:
\begin{itemize}
\item
This study provides the first systematic analysis of code generation, execution, and self-debugging's role in mathematical problem-solving. Our findings reveal that \gptcode's impressive mathematical problem-solving proficiency is primarily attributed to its step-by-step code generation and dynamic solution refinement based on code execution outcomes.

\item
We introduce the innovative \textit{explicit code-based self-verification (CSV) prompt}, which leverages \gptcode's advanced code generation mechanism. This prompt guides the model to verify the answer and then reevaluate its solution with code. CSV not only extends the verification to the logic behind problem-solving but also improves the efficacy of the majority voting method by integrating the verification states.
\item 
{Additionally, we have contributed to the LLM community by creating two new instruction-following datasets: MATH-code and MMLU-Math-code. These datasets are designed to enhance the mathematical reasoning capabilities of open-source models.}
\end{itemize}

\begin{figure}[t]
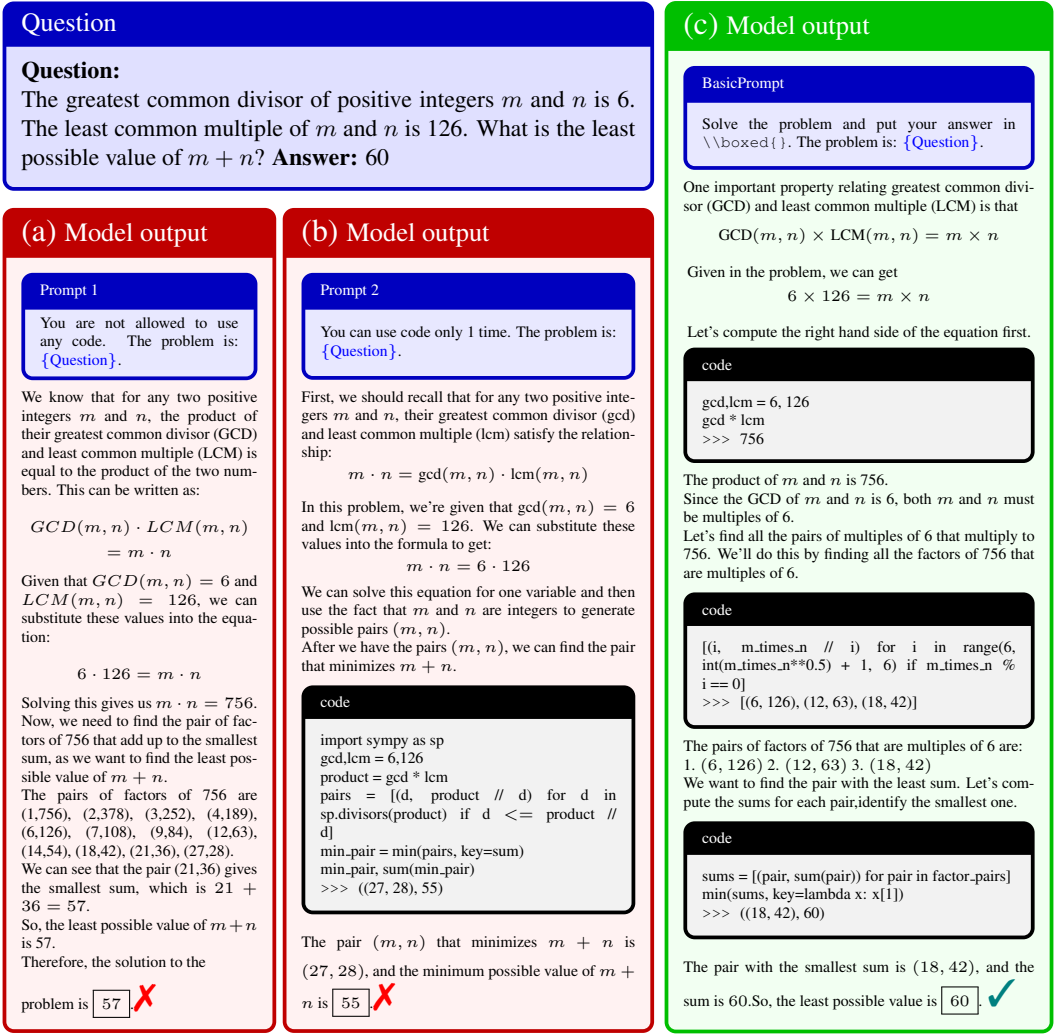


\newlength{\oldcolumnsep}
\setlength{\oldcolumnsep}{\columnsep}
\setlength{\columnsep}{3pt}

\noindent 
\begin{minipage}{0.62\textwidth}
\begin{tcolorbox}[colback=blue!12!white,colframe=blue!75!black,title=Question,
    left=1mm,  
    right=1mm, 
    top=1mm,   
    bottom=1mm 
]
\begin{small}
\textbf{Question:}\\
The greatest common divisor of positive integers $m$ and $n$ is 6. The least common multiple of $m$ and $n$ is 126. What is the least possible value of $m+n$? \textbf{Answer:} 60
\end{small}

\vspace{0.2mm}
\end{tcolorbox}

\begin{minipage}{0.42\textwidth}
\begin{tcolorbox}[colback=red!5!white,colframe=red!75!black,title=\large(a)~\normalsize Model output,
    left=1mm,  
    right=1mm, 
    top=1mm,   
    bottom=1mm 
]
\begin{tiny}
\begin{tcolorbox}[colback=blue!9!white,colframe=blue!75!black,title=Prompt 1,
    left=1mm,  
    right=1mm, 
    top=1mm,   
    bottom=1mm 
]
\vspace{-1mm}
You are not allowed to use any code. The problem is: \textcolor{blue}{\{Question\}}.
\vspace{-1mm}
\end{tcolorbox}
We know that for any two positive integers \(m\) and \(n\), the product of their greatest common divisor (GCD) and least common multiple (LCM) is equal to the product of the two numbers. This can be written as:
\[
    GCD(m, n) \cdot LCM(m, n)
  \] 
\vspace{-4mm}
  \[ 
     =m \cdot n
\]

Given that \(GCD(m, n) = 6\) and \(LCM(m, n) = 126\), we can substitute these values into the equation:

\[6 \cdot 126 = m \cdot n\]

Solving this gives us \(m \cdot n = 756\).

Now, we need to find the pair of factors of 756 that add up to the smallest sum, as we want to find the least possible value of \(m+n\).

The pairs of factors of 756 are (1,756), (2,378), (3,252), (4,189), (6,126), (7,108), (9,84), (12,63), (14,54), (18,42), (21,36), (27,28). 

We can see that the pair (21,36) gives the smallest sum, which is \(21+36=57\).

So, the least possible value of \(m+n\) is 57.

Therefore, the solution to the 

problem is \(\boxed{57}\).\Large \textcolor{red}{\xmark}

\end{tiny}
\vspace{-0.5mm}
\end{tcolorbox}
\end{minipage}
\hfill
\begin{minipage}{0.57\textwidth}
\begin{tcolorbox}[colback=red!5!white,colframe=red!75!black,title=\large(b)~\normalsize Model output,
    left=1mm,  
    right=1mm, 
    top=1mm,   
    bottom=1mm 
]
\begin{tiny}
\begin{tcolorbox}[colback=blue!9!white,colframe=blue!75!black,title=Prompt 2,
    left=1mm,  
    right=1mm, 
    top=1mm,   
    bottom=1mm 
]
\vspace{0.2mm}
You can use code only 1 time. The problem is: \textcolor{blue}{\{Question\}}.
\vspace{0.2mm}
\end{tcolorbox}
First, we should recall that for any two positive integers \(m\) and \(n\), their greatest common divisor (gcd) and least common multiple (lcm) satisfy the relationship:
\vspace{-2mm}
\[
m \cdot n = \text{{gcd}}(m, n) \cdot \text{{lcm}}(m, n)
\]
\vspace{-3mm}

In this problem, we're given that \(\text{{gcd}}(m, n) = 6\) and \(\text{{lcm}}(m, n) = 126\). We can substitute these values into the formula to get:
\vspace{-2mm}
\[
m \cdot n = 6 \cdot 126
\]
\vspace{-4mm}

We can solve this equation for one variable and then use the fact that \(m\) and \(n\) are integers to generate possible pairs \((m, n)\).

After we have the pairs \((m, n)\), we can find the pair that minimizes \(m+n\).
\begin{tcolorbox}[colback=black!5!white,colframe=black!75!black,title=code,
    left=1mm,  
    right=1mm, 
    top=1mm,   
    bottom=1mm 
]
import sympy as sp\\
gcd,lcm = 6,126\\
product = gcd * lcm\\
pairs = [(d, product // d) for d in sp.divisors(product) if d $<=$ product // d]\\
min\_pair = min(pairs, key=sum)\\
min\_pair, sum(min\_pair)\\
\texttt{>>> }((27, 28), 55)
\end{tcolorbox}
The pair \((m, n)\) that minimizes \(m+n\) is \((27, 28)\), and the minimum possible value of \(m+n\) is \(\boxed{55}\).\Large\textcolor{red}{\xmark}
\end{tiny}
\vspace{-0.36mm}
\end{tcolorbox}
\end{minipage}
\end{minipage}
\hfill
\begin{minipage}{0.37\textwidth}
\begin{tcolorbox}[colback=green!5!white,colframe=green!75!black,title=\large(c)~\normalsize Model output,
    left=1mm,  
    right=1mm, 
    top=1mm,   
    bottom=1mm 
]
\begin{tiny}
\begin{tcolorbox}[colback=blue!9!white,colframe=blue!75!black,title=BasicPrompt,
    left=1mm,  
    right=1mm, 
    top=1mm,   
    bottom=1mm 
]
Solve the problem and put your answer in \verb|\\boxed{}|. The problem is: \textcolor{blue}{\{Question\}}.
\end{tcolorbox}
One important property relating greatest common divisor (GCD) and least common multiple (LCM) is that
\vspace{-1mm}
\[
\text{{GCD}}(m, n) \times \text{{LCM}}(m, n) = m \times n
\]
\vspace{-0.79mm}
Given in the problem, we can get 
\vspace{-1mm}
\[
6 \times 126 = m \times n
\]
\vspace{-0.62mm}
Let's compute the right hand side of the equation first.
\begin{tcolorbox}[colback=black!5!white,colframe=black!75!black,title=code,
    left=1mm,  
    right=1mm, 
    top=1mm,   
    bottom=1mm 
]
gcd,lcm = 6, 126\\
gcd * lcm\\
\texttt{>>> }{756}
\end{tcolorbox}
The product of \(m\) and \(n\) is 756.

Since the GCD of \(m\) and \(n\) is 6, both \(m\) and \(n\) must be multiples of 6.

Let's find all the pairs of multiples of 6 that multiply to 756. We'll do this by finding all the factors of 756 that are multiples of 6.

\begin{tcolorbox}[colback=black!5!white,colframe=black!75!black,title=code,
    left=1mm,  
    right=1mm, 
    top=1mm,   
    bottom=1mm 
]
[(i, m\_times\_n // i) for i in range(6, int(m\_times\_n**0.5) + 1, 6) if m\_times\_n \% i == 0]\\
\texttt{>>> }[(6, 126), (12, 63), (18, 42)]
\end{tcolorbox}
The pairs of factors of 756 that are multiples of 6 are:

1. \( (6, 126) \)
2. \( (12, 63) \)
3. \( (18, 42) \)

We want to find the pair with the least sum. Let's compute the sums for each pair,identify the smallest one.
\begin{tcolorbox}[colback=black!5!white,colframe=black!75!black,title=code,
    left=1mm,  
    right=1mm, 
    top=1mm,   
    bottom=1mm 
]
sums = [(pair, sum(pair)) for pair in factor\_pairs]\\
min(sums, key=lambda x: x[1])\\
\texttt{>>> }((18, 42), 60)

\end{tcolorbox}
The pair with the smallest sum is \( (18, 42) \), and the sum is \( 60 \).So, the least possible value is \(\boxed{60}\).
\Large \textcolor{teal}{\cmark}
\end{tiny}
\end{tcolorbox}
\end{minipage}

\setlength{\columnsep}{\oldcolumnsep}
\caption{Example from the 277th number theory problem in the MATH dataset: \textbf{(a)} Prompt~1~generates intermediate text-only reasoning steps. \textbf{(b)} Prompt~2~generates intermediate text reasoning steps and post-hoc Python code, akin to PAL~\citep{gao2023pal}. \textbf{(c)} Basic prompt that set no constrain on code usage. Details of the three prompts are presented in Sec.~\ref{pilot-experiments}. (For more examples, see Tab.~\ref{tab:code_freq_example1} and Tab.~\ref{tab:code_freq_example2}.) 
} 
\label{fig:example1}
\end{figure}

\section{Related work}

\textbf{Chain-of-Thought Reasoning.}
The \textit{Chain-of-Thought (CoT)} prompting approach proposed by~\citet{wei2022chain} is a notable contribution that showcases the multi-step reasoning capabilities of LLMs.
By simply adding “Let's think step by step" before questions,~\cite{NEURIPS2022_8bb0d291} implements \textit{Zero-shot-CoT}, which can serve as a strong zero-shot baseline. Further research extends the reasoning capabilities of CoT by applying majority voting to improve self-consistency~\citep{wang2023selfconsistency}, choosing few-shot examples and output chains with more complex reasoning steps~\citep{fu2022complexity}, breaking down the problem to simpler sub-problems~\citep{zhou2023leasttomost}, or even expanding Chain-of-thought to Tree-of-Thoughts~\citep{Yao2023TreeOT}. Similar to Zero-shot-CoT, our method apply  “step by step"-like prompts to regularize \gptcode's use of code without the careful design of step-by-step few-shot examples. Additionally, We enhance majority voting to verification-guided weighted majority voting, leveraging the results of CSV as voting weights.

\textbf{Solving Math Problems with Code.}
Large language models have been found to be less accurate in performing arithmetic calculations, such as addition, subtraction, multiplication, etc
\citep{cobbe2021gsm8k, lewkowycz2022solving, gao2023pal, lu2022dynamic}. Consequently, previous works have attempted to solve math problems with the assistance of code. The GSM8K dataset~\citep{cobbe2021gsm8k} uses calculation annotations to extract all arithmetic calculations solved by an external calculator: the Python \textit{eval} function. To further leverage the role of code in LLMs, \textit{Program-Aided Language model (PAL)}~\citep{gao2023pal} as well as \textit{Program of Thoughts (PoT)}~\citep{chen2022program} interpret the math problems as Python codes and execute the codes with an external Python interpreter to obtain the answer. 
Although they can get more accurate answers than some non-code methods, many generated codes have execution errors or get wrong answers due to the lack of verification mechanism. Our approach not only utilizes the ability of \gptcode~to generate multi-step codes and refine codes that fail to run, but also uses CSV to enhance the reliability and accuracy of the answers.

\textbf{Self-Verification.} 
Human problem solving is not always a one-time success, but rather requires iterative thinking, verification, and refinement. Unlike previous studies that train an additional verifier to verify the correctness of final answers~\citep{cobbe2021gsm8k} or intermediate steps~\citep{lightman2023lets, li2023making}, ~\citet{weng2023large} showed the self-verification abilities of LLMs by generating multiple answers and ranking them by self-verification scores. Furthermore, \textit{SELF-REFINE} proposed by~\citet{madaan2023self} iteratively refines its output through self-generated feedback. Unlike these self-verification methods that require LLMs to give verification feedback in natural language, our method applies generated codes to verify the answers and votes on different answers based on the verification results, thus improving the accuracy of the verification and making full use of the information in the verification process.

\section{Method}

We first conduct a pilot experiment with \gptcode ~on the challenging MATH dataset~\citep{hendrycks2021measuring}. Remarkably, it achieves an accuracy of 69.7\%, significantly 
surpassing the previous state-of-the-art performance of 53.9\%~\citep{zheng2023progressive}. Encouraged by the compelling performance of \gptcode, we strive to systematically explore and analyze its underlying code mechanisms. In Sec.~\ref{pilot-experiments}, we illustrate, via our code-constrained prompts design, that \gptcode's robust performance in solving 
math problems derives not only from its ability to generate accurate step-by-step code, but also from its self-debugging mechanism. In Sec.~\ref{proposed method}, we aim to leverage \gptcode's self-debugging strengths to further improve its mathematical problem-solving ability.

\subsection{Pilot Experiments on analyzing Code Usage of \gptcode}
\label{pilot-experiments}

To explore the impact of code usage on \gptcode 's math problem-solving capabilities, we adopt a straightforward approach by constraining \gptcode's interaction with code through thoughtfully constructed prompts. Specifically, we introduce two code-constrained prompts and the basic prompt for comparison:

\begin{itemize}
\item 
\textbf{Prompt 1: } \textit{No code usage is allowed}: In response to this prompt, \gptcode~is prohibited from incorporating code into its solutions. This prompts \gptcode~to rely solely on Natural Language (\textbf{NL}) reasoning chain, resembling solutions in the CoT framework~\citep{wei2022chain}. The resulting sequence of reasoning steps is depicted as \(\mathbf{C_{\scriptscriptstyle{NL}}}\), with an example given in Fig.~\ref{fig:example1} (a).

\item
\textbf{Prompt 2:} \textit{Code can be used only once}: In this prompt setting, \gptcode~is permitted to employ code exclusively within a single code block to generate the solution, mirroring the PAL approach introduced by~\citet{gao2023pal}. We denote this sequence as \(\mathbf{C_{\scriptscriptstyle{SL}}}\), representing a series of Symbolic Language (\textbf{SL}), such as Python. An example is shown in Fig.~\ref{fig:example1}~(b).

\item 
\textbf{Basic Prompt:} \gptcode~is prompted to tackle the problem without any restrictions on code usage. This prompt leads to \gptcode's typical functioning pattern, which can be denoted as $\mathbf{C = ((c1_{\scriptscriptstyle{NL}},c1_{\scriptscriptstyle{SL}}),(c2_{\scriptscriptstyle{NL}},c2_{\scriptscriptstyle{SL}}),\dots)}$, representing a sequential list of reasoning steps, each consisted of both natural language and Python code, with an example shown in Fig.~\ref{fig:example1} (c). 

\end{itemize}

\begin{figure}[t]
\begin{center}
\includegraphics[width=1\textwidth]{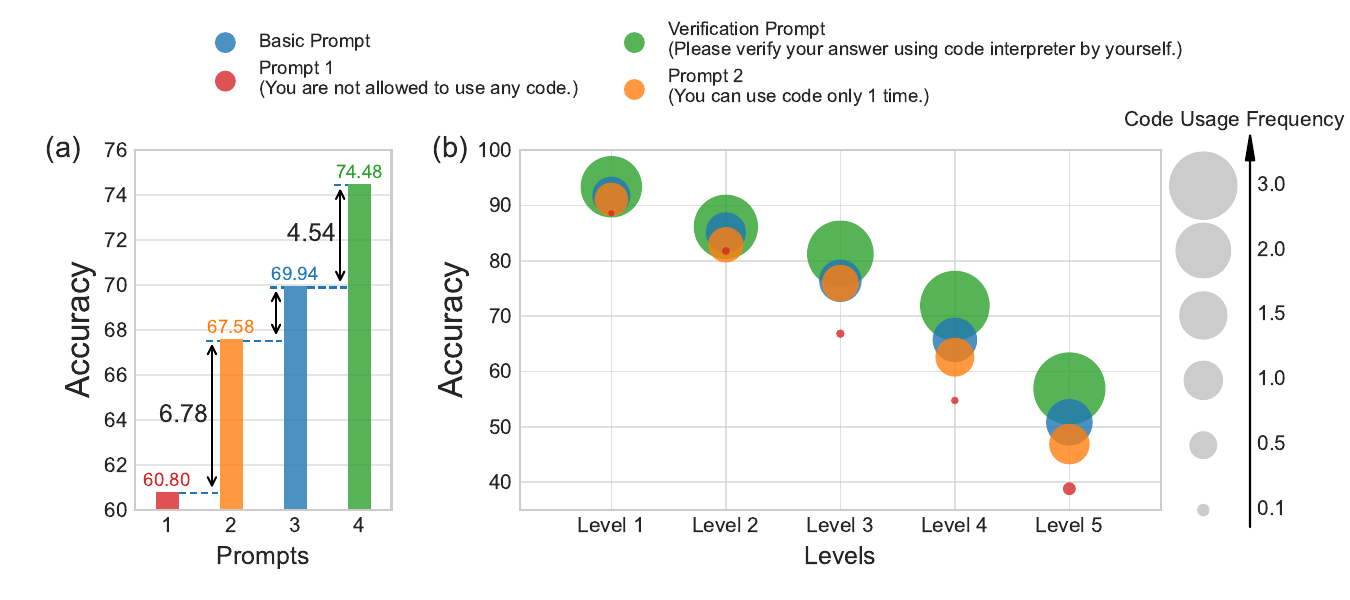}

\end{center}
\vspace{-7mm}
\caption{\small
Performance on MATH dataset of different levels by applying different prompts to adjust the frequency of code usage. \textbf{(a)} Comparison of overall accuracy between the 4 prompts. \textbf{(b)} Code usage frequency is in proportion to accuracy in all five levels and this phenomenon is especially apparent when the problems are relatively complicated ~(i.e. with higher level).
}
\label{fig:pilot_experiments}
\end{figure}

Apart from the specific example in Fig.~\ref{fig:example1}, we introduce \textit{Code Usage Frequency} to record the number of the code execution for different prompts. The results of the experiments using these prompts are shown in Fig.~\ref{fig:pilot_experiments} (b). This figure illustrates a positive correlation between the better performance of \gptcode~and the higher Code Usage Frequency. More specifically,

\textbf{Prompt 1} v.s. \textbf{Prompt 2:} \textit{Prompt 1} 
 results in almost negligible code usage, while \textit{Prompt 2} results in approximately 1 time's 
code usage. \textit{Prompt 2} yields an accuracy gain of 6.9 percent over Prompt 1. This suggests that the Python code chains $\mathbf{C_{SL}}$, can improve computational capability more than the natural language chains $\mathbf{C_{NL}}$. This observation is consistent with the findings in previous Python-based prompting methods~\citep{gao2023pal, chen2022program}. However, employing code only once comes with an inherent drawback – the model lacks the ability to self-debugging when the code output triggers an error or produces an implausible outcome.

\textbf{Prompt 2} v.s.~\textbf{Basic Prompt}: The \textit{Basic Prompt} consistently produces solutions that entail multiple instances of code usage, resulting in a large Code Usage Frequency. Additionally, the \textit{Basic Prompt} exhibits notably enhanced accuracy. These improvements in Code Usage Frequency and accuracy might be attributable to two unique advantages: (1) Generating code in brief and frequent segments, divided among natural language reasoning steps, tends to result in higher accuracy. (2) The model possesses the capability to evaluate the results of code execution and make corrections to solution steps if the outcomes contain bugs or are deemed illogical, as illustrated in Tab.~\ref{tab:baseline_example1} and Tab.~\ref{tab:baseline_example2}.

From these observations, it is plausible to enhance and build upon the favorable attributes of \gptcode,  to further improve its precision in tackling math problems.

\subsection{Explicit Code-based Self-verification Prompting}
\label{proposed method}


Inspired by the observations on Code Usage Frequency analysis, we seek to harness the capabilities of \gptcode. These capabilities include the model's aptitude for generating accurate code, evaluating the outcomes of code execution, and automatically adjusting reasoning steps of solutions when needed. However, despite these advantages, \gptcode~currently falls short in assuring solution correctness. Consequently, our objective is to utilize these strengths to augment solution verification.

\setlength{\columnsep}{1cm}
\begin{figure}[t]
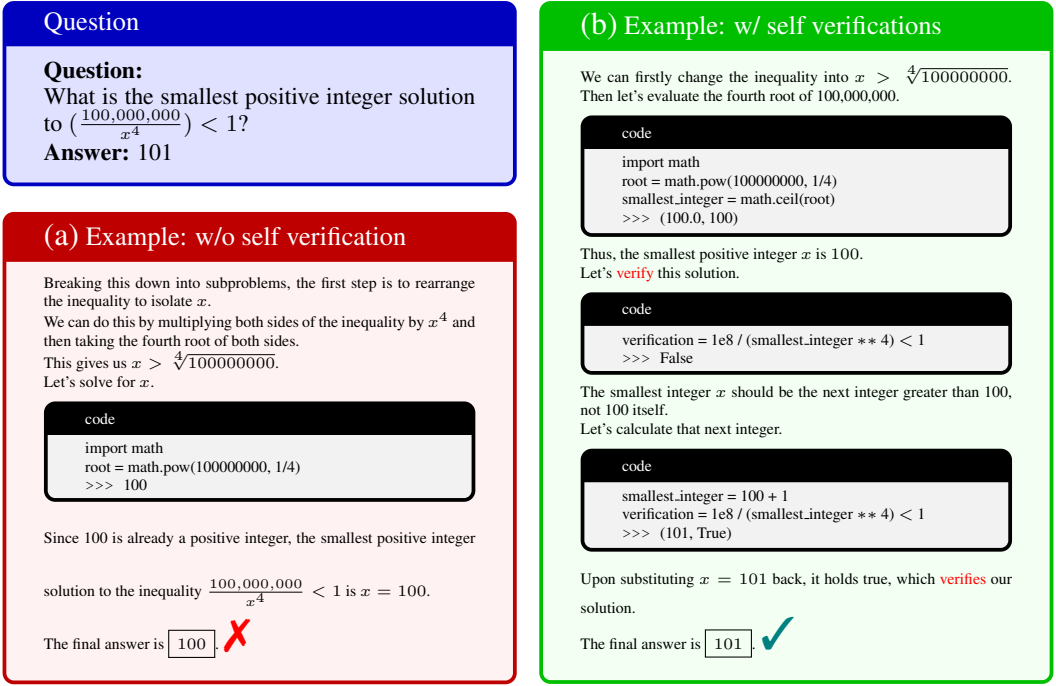

\setlength{\columnsep}{3mm}
\begin{multicols}{2}
\begin{tcolorbox}[colback=blue!12!white,colframe=blue!75!black,title=Question]
\vspace{-1mm}
\begin{small}
\textbf{Question:}\\
What is the smallest positive integer solution to $(\frac{100,000,000}{x^4}) < 1$?

\textbf{Answer:}
101
\end{small}
\end{tcolorbox}
\vspace{-1.5mm}
\begin{tcolorbox}[colback=red!5!white,colframe=red!75!black,title=\large(a)\normalsize~Example: w/o self verification]
\vspace{-1mm}
\begin{tiny}
Breaking this down into subproblems, the first step is to rearrange the inequality to isolate $x$.\\
We can do this by multiplying both sides of the inequality by $x^4$ and then taking the fourth root of both sides.\\
This gives us $x>\sqrt[4]{100000000}$.

Let's solve for $x$.
\begin{tcolorbox}[colback=black!5!white,colframe=black!75!black,title=code]
\vspace{-2mm}

\begin{tiny}
import math\\
root = math.pow(100000000, 1/4)
\end{tiny}

\texttt{>>> }100
\vspace{-2mm}
\end{tcolorbox}
\vspace{-2mm}

Since 100 is already a positive integer, the smallest positive integer solution to the inequality $\frac{100,000,000}{x^4} < 1$ is $x = 100$.\\
The final answer is \boxed{100}. \LARGE \textcolor{red}{\xmark}

\end{tiny}
\end{tcolorbox}

\begin{tcolorbox}[colback=green!5!white,colframe=green!75!black,title=\large(b)\normalsize~Example: w/ self verifications]
\vspace{-1mm}
\begin{tiny}
We can firstly change the inequality into $x>\sqrt[4]{100000000}$. Then let's evaluate the fourth root of 100,000,000.

\begin{tcolorbox}[colback=black!5!white,colframe=black!75!black,title=code]
\vspace{-2mm}

\begin{tiny}
import math\\
root = math.pow(100000000, 1/4)\\
smallest\_integer = math.ceil(root)
\end{tiny}

\texttt{>>> }(100.0, 100)
\vspace{-2mm}
\end{tcolorbox}
Thus, the smallest positive integer $x$ is $100$.\\
Let's \textcolor{red}{verify} this solution.
\begin{tcolorbox}[colback=black!5!white,colframe=black!75!black,title=code]
\vspace{-2mm}

\begin{tiny}
verification = 1e8 / (smallest\_integer $\ast\ast$ 4) $<$ 1
\end{tiny}

\texttt{>>> }False
\vspace{-2mm}
\end{tcolorbox}
The smallest integer $x$ should be the next integer greater than 100, not 100 itself.\\
Let's calculate that next integer.
\begin{tcolorbox}[colback=black!5!white,colframe=black!75!black,title=code]
\vspace{-2mm}

\begin{tiny}
smallest\_integer = 100 + 1\\
verification = 1e8 / (smallest\_integer $\ast\ast$ 4) $<$ 1\\
\end{tiny}
\texttt{>>> }(101, True)
\vspace{-2mm}
\end{tcolorbox}
Upon substituting $x=101$ back, it holds true, which \textcolor{red}{verifies} our solution.\\
The final answer is \boxed{101}. \LARGE \textcolor{teal}{\cmark}
\end{tiny}
\end{tcolorbox}
\end{multicols}
\vspace{-5mm}
\caption{\small
 Question from the 712th intermediate algebra problem in the MATH dataset. \textbf{(a)} Without self-verification, the model generates a wrong answer. \textbf{(b)} With self-verification, the model corrects the error and generates the correct answer. The CSV prompt: \textit{To solve the problem using code interpreter step by step, and please verify your answer using code interpreter. }
}
\label{fig:examples2}
\end{figure}

To achieve this objective, we propose the technique termed as \textbf{explicit code-based self-verification} (CSV). This method prompts \gptcode~to explicitly validate its answer through code generation. By implementing this prompt, we introduce an extra verification stage to the solution $\mathbf{C}$, referred to as $\mathbf{V}$. The verification outcome $\mathbf{V}$ can be classified as \textit{True}, \textit{False}, or \textit{Uncertain}. An \textit{Uncertain} classification indicates that \gptcode~encountered difficulties in identifying an effective method for answer verification, thereby abstaining from delivering a definitive verification result. Leveraging \gptcode's inherent autonomous capabilities, we can formulate the proposed prompting as follows:

\[
\mathbf{C} \rightarrow \mathbf{V}=\begin{cases}
    \text{True} & \rightarrow \text{final answer} \\
    \text{False} & \rightarrow \mathbf{C}_{\text{new}} \rightarrow \mathbf{V} \rightarrow \dots \rightarrow \text{True} \rightarrow \text{final answer} \\
    \text{Uncertain} & \rightarrow \text{final answer}
\end{cases}
\]

An example is presented in Fig.~\ref{fig:examples2} (b). Incorporated with CSV, the model becomes capable of using code to verify answers, then reviewing and adjusting how it arrived at the solution if the verification result is \textit{False}, aiming at obtaining the correct answer. Upon refining and correcting the initial solution, we anticipate a notable increase in accuracy. It is worth noting that both the verification and rectification stages are code-based. This inevitably results in increased Code Usage Frequency, akin to the aforementioned analysis, which will be further demonstrated in subsequent experiments.

We perform experiments with CSV, and these results can be found in Fig.~\ref{fig:pilot_experiments}. The
experiment here is conducted with \gptcode~on MATH \citep{hendrycks2021measuring}. In Fig.~\ref{fig:pilot_experiments} (b), the accuracy achieved with our proposed CSV prompt consistently surpasses that of the \textit{Basic Prompt} across all designated difficulty levels\footnote{Human-perceived easier problems are categorized under Level-1 difficulty as per \cite{hendrycks2021measuring}.}. Meanwhile, the Code Usage Frequency receives a clear increase.

Before the advent of \gptcode, prior frameworks~\citep{lightman2023lets,cobbe2021gsm8k} depended on an external LLM to use natural language for verification and well-designed few-shot example prompts. In contrast, our approach simplifies the process by relying solely on a straightforward prompt for \gptcode, all in a \textbf{zero-shot} manner. This enables \gptcode~to autonomously verify and independently rectify its solutions using the advanced code execution mechanism, thereby eliminating the need for customized few-shot examples.

Given that CSV can effectively verify problem-solving answers, we can naturally integrate the verification states into majority voting, akin to the methodology embraced in self-consistency CoT~\citep{wang2023selfconsistency}. Answers deemed \textbf{True} through verification are generally more trustworthy, reflecting the problem-solving approach seen in human cognition~\citep{newell1972human, WANG201081}. This improved reliability can be leveraged in the widely-used majority voting process. To exploit this insight, we introduce \textit{verification-guided weighted majority voting}, which assigns different weights to the states of the verification process.

In practice, it sometimes occurs that once an answer is confirmed as False, no additional verification is conducted, yielding a False verification state. We allocate corresponding weights these states of \textbf{True}, \textbf{Uncertain}, \textbf{False}: $w_\mathbf{\scriptscriptstyle{T}}, w_\mathbf{{\scriptscriptstyle{U}}}$, and $w_\mathbf{{\scriptscriptstyle{F}}}$, respectively.

\begin{figure}[t]
    \centering
    \includegraphics[width=0.99\textwidth]{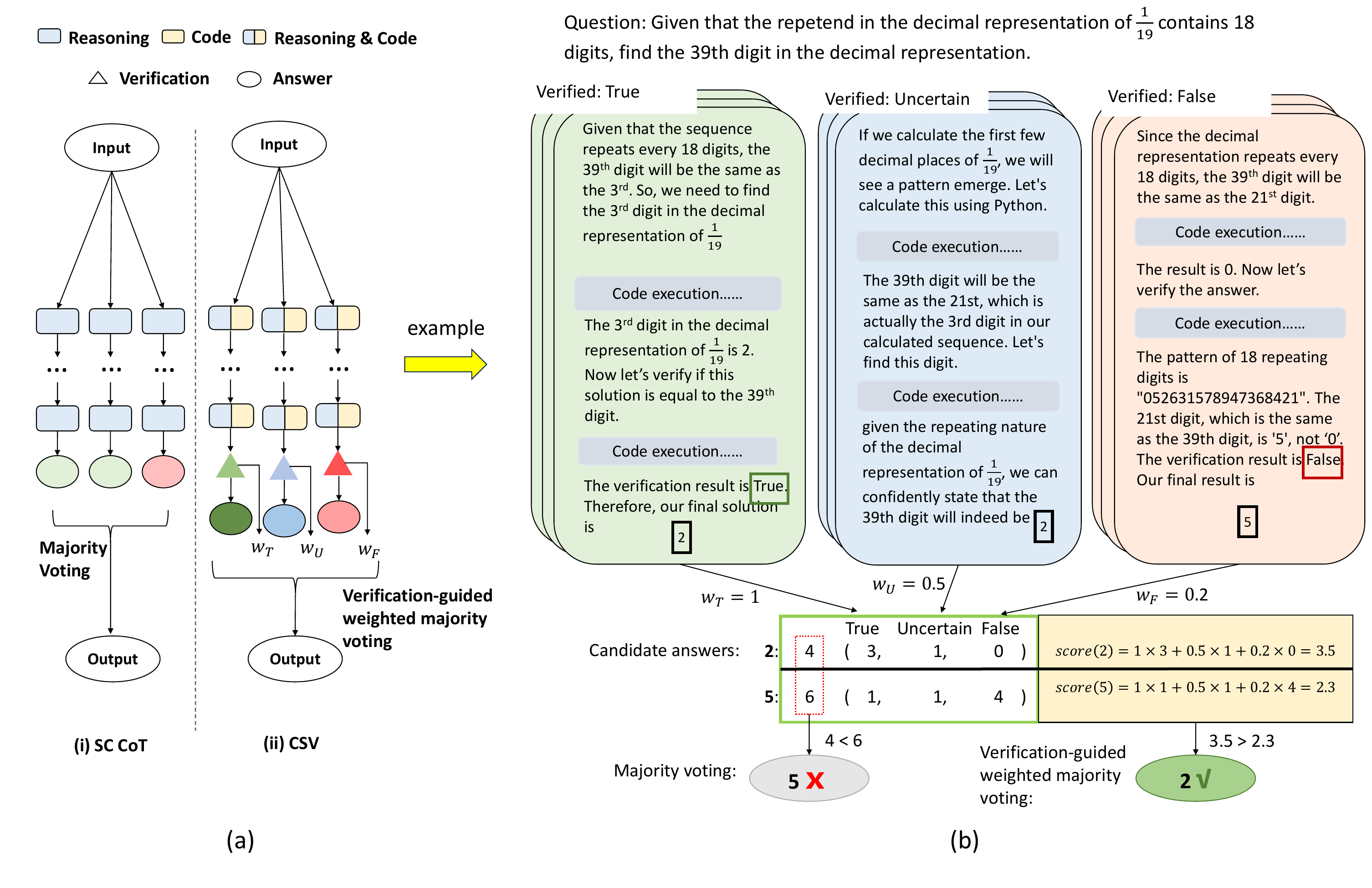}
    \vspace{-4mm}
    \caption{\textbf{(a)} Illustration of the Naive majority voting~\citep{wang2023selfconsistency} and our Verification-guided weighted majority voting. The full pipeline of the proposed Verification-guided Weighted Majority Voting framework. We use the model to generate several different solutions. Then we detect the self-verification state of each solution, and classify them into three states: \textit{True}, \textit{Uncertain}, and \textit{False}. According to the state of the verification, we assign each solution a different weight, and use the classified result to vote the score of each possible answer. }
    
\label{fig:combined}
\end{figure}

Similar to the {Self-consistency with CoT (CoT-SC)}~\citep{wang2023selfconsistency} in Fig.~\ref{fig:combined} (a)(ii), our framework can sample $k$ paths. For simplicity, we extract pairs of final answers and their corresponding verification results from $k$ solutions, represented as $(v^i, a^i), i=1,2,\dots ,k$, where $v^i$ and $a^i$ denote the $i$-th final answer and final verification result, respectively. 


So the voting score for each candidate answer \(a\) can be expressed as:

\begin{equation}
\label{eq1}
\text{Score}(a) = \sum_{\{v^i\}} w_v (\#\{i \mid  a^{i}=a~ \text{and} ~ v^{i}=v\}), \quad v\in\{\text{True}, \text{Uncertain}, \text{False}\},
\end{equation}

Here, $a$ represents a candidate answer, $v$ denotes the state of verification, and $w_v$ is an element from the set $\{w_\mathbf{\scriptscriptstyle{T}}, w_\mathbf{{\scriptscriptstyle{U}}}, w_\mathbf{{\scriptscriptstyle{F}}}\}$. Each $w_v$ signifies the degree of confidence associated with its corresponding verification state.

Finally, we select the answer with the highest score from all candidate answers,

\begin{equation}
    \text{Output} = \mathop{\arg\max}\limits_{a} \text{Score}(a) ,
\end{equation}

where $\text{Score}(a)$ refers to the score of answer $a$ according to Eq.~\ref{eq1}.

It should be noted that when $w_v = 1$ for all $w_v \in \{w_\mathbf{\scriptscriptstyle{T}}, w_\mathbf{{\scriptscriptstyle{U}}}, w_\mathbf{{\scriptscriptstyle{F}}}\}$, Eq.~\ref{eq1} becomes equivalent to the naive majority voting employed in Self-Consistency with CoT (CoT-SC)~\citep{wang2023selfconsistency}. Typically, we set ${w_\mathbf{\scriptscriptstyle{T}} > w_\mathbf{{\scriptscriptstyle{U}}} > w_\mathbf{{\scriptscriptstyle{F}}}}$, which means that an answer verified true has a greater confidence than the one with an uncertain state of verification, while an answer verified false has the lowest degree of confidence. An example of the calculation process within verification-guided weighted majority voting is illustrated in Fig.~\ref{fig:combined}. 

\section{Experiments}

\subsection{Performance on MATH}

The MATH dataset~\citep{hendrycks2021measuring} is recognized as the most challenging math word problem dataset, as also highlighted by Chen et al.~\citep{chen2023theoremqa}. Most of our experiments and the corresponding analyses are performed on the MATH benchmark. Tab.~\ref{tab:gpt4_math} compares the performance of the \gptcode~against other models. \gptcode~reaches 69.69\% on MATH~\citep{Hendrycks2020MeasuringMM}, largely surpassing the previous state of art result (53.90\%), which shows that \gptcode~exhibits strong abilities in solving math problems and is used as our baseline for ablation study. On top of \gptcode, our method further improves its accuracy, raising the result to 73.54\% after adding explicit code-based self-verification, and 84.32\% after adding both explicit code-based self-verification and verification-guided
weighted majority voting (the number of sampled paths is 16). Note that this astonishingly high result is based on the strong abilities of the base model \gptcode, and our method amplifies its good qualities of \gptcode, with the ability to verify solutions.

Note that although adding Code-based Self-verification can improve the performance of every individual subject, the extent of improvement varies from subject to subject, from 7.6\% to only 0.6\%. In particular, the Geometry problem only has an increased accuracy of 0.6\%, even though the original \gptcode~accuracy is only 54.0\%, which is low among the subjects. This discrepancy may be attributed to the fact that solving geometry problems often requires multi-modality~\citep{chen2023theoremqa}, a concept beyond the scope of this paper.

\definecolor{c1}{HTML}{DC143C}
\definecolor{c2}{HTML}{32CD32}

\begin{table}[t]
   \caption{Accuracy (\%) on MATH dataset. \textbf{VW-voting} is the abbreviation for the verification-guided weighted majority voting. (\textbf{Overall:} The results across various MATH subtopics~\citep{hendrycks2021measuring}) 
   }
  
   \label{tab:gpt4_math}
   \centering
    \resizebox{\columnwidth}{!}{
   \begin{tabular}{lcccccccccc}
     \toprule
     
 & \textbf{Code-based}  & \textbf{VW-} & Intermediate & Precalculus &Geometry & Number &Counting \& &PreAlgebra  & Algebra &Overall  \\
 &\textbf{Verification} & \textbf{Voting} &Algebra & \textbf{--}  & \textbf{--} & Theory & Probability& \textbf{--}  & \textbf{--} & MATH\\
     \cmidrule(r){2-2} \cmidrule(r){3-3} \cmidrule(r){4-11}  
 GPT-4  & \xmark& \xmark & - & - & -&-&-&-& - &42.20\\
 \cmidrule(r){1-11} 
 {\shortstack{GPT-3.5}}&\xmark& \xmark  &14.6   & 16.8& 22.3&33.4 &29.7 & 53.8&49.1& 34.12\\
 {\shortstack{GPT-4} (CoT)}&\xmark&\xmark  &23.4   & 26.7& 36.5&49.6 &53.1 & 71.6&70.8& 50.36\\
 {\shortstack{GPT-4 (PHP)}} &\xmark&\xmark  &26.3   & 29.8& 41.9&55.7 &56.3 &73.8&74.3& 53.90\\
 \cmidrule(r){1-11} 
 {\shortstack{\gptcode~}}&\xmark&\xmark  & 50.1&51.5&53.4&77.2&70.6&86.3&83.6&69.69 \\
 \cmidrule(r){1-11}
 {\shortstack{\gptcode}~+ \textbf{CSV}}&\checkmark&\xmark  &56.6&53.9&54.0&85.6&77.3&86.5&86.9&73.54 \\
 \textcolor{c1}{\textit{Improvement}} &&& \textcolor{c1}{\textbf{ +6.5}} & \textcolor{c1}{\textbf{ +2.4}} & \textcolor{c1}{\textbf{ +0.6}} & \textcolor{c1}{\textbf{ +7.6}} & \textcolor{c1}{\textbf{ +6.7}} & \textcolor{c1}{\textbf{ +0.2}} & \textcolor{c1}{\textbf{ +3.3}} & \textcolor{c1}{\textbf{ +3.85}} \\
 {\shortstack{\gptcode}~+ \textbf{CSV} + \textbf{Voting}}&\checkmark&\checkmark\textbf{ (k=16)}&\textbf{74.4}&\textbf{67.8}&\textbf{64.9}&\textbf{94.1}&\textbf{89.0}&\textbf{91.6}&  \textbf{95.6}&\textbf{84.32}\\   

 \textcolor{c1}{\textit{Improvement}} & &  &\textcolor{c1}{\textbf{ +24.3}}&\textcolor{c1}{\textbf{ +16.3}}&\textcolor{c1}{\textbf{ +11.5}}&\textcolor{c1}{\textbf{ +16.9}}&\textcolor{c1}{\textbf{ +18.4}}&\textcolor{c1}{\textbf{ +5.3}}&\textcolor{c1}{\textbf{ +12.0}}&\textcolor{c1}{\textbf{ +14.63}}\\
     \bottomrule
   \end{tabular}}
 \end{table}

\subsection{Performance on other datasets}

In addition to the challenging MATH dataset, we have also performed our method on other reasoning datasets such as GSM8K~\citep{cobbe2021gsm8k}, MMLU-Math, and MMLU-STEM~\citep{Hendrycks2020MeasuringMM}. The corresponding results can be viewed in Tab.~\ref{tab:gsm8k} and Tab.~\ref{tab:mmlu_math}. When integrated on top of GPT-4-code, our method outperforms other methods in the competition, achieving state-of-the-art results across all datasets. Other subjects in MMLU benchmarks are provided in Fig.~\ref{fig:mmlu}.  A comparative analysis of our results with those of previous state-of-the-art techniques and open-source models is also provided. 

Tab.~\ref{tab:gsm8k} illustrates that verification-guided majority voting is an effective framework to reduce the number of sampled paths, compared to GPT-4 with model~selection~\citep{zhao2023automatic} and PHP~\citep{zheng2023progressive}.

Tab.~\ref{tab:mmlu_math} presents a comparison of our model's performance with existing models~\citep{hoffmann2022training,taylor2022galactica} on the MMLU-Math dataset and with state-of-the-art open-sourced models\footnote{\url{https://huggingface.co/spaces/HuggingFaceH4/open_llm_leaderboard}} on MMLU-STEM. The open-source models remain significantly outpaced by their closed-source counterparts. To address this gap,  we have made the dataset and will make it publicly available in the near future. Our intention is to facilitate the fine-tuning of open-source LLMs. For example, the open-source model LLaMA 2~\citep{touvron2023llama} can potentially utilize this data to further bolster its math reasoning capabilities.

\begin{table}[t]
    \footnotesize
	\centering
	\begin{minipage}[t]{0.48\textwidth}
	\centering
	\caption{Performance on GSM8K dataset.}
	\resizebox{0.92\columnwidth}{!}{
	\begin{tabular}[b]{ lccc }
		\toprule
		 Method & Sampled paths  & Accuracy(\%)  \\
		\midrule
            GPT-3.5 (5-shot) &-- & 57.1    \\
            GPT-4 (5-shot CoT) &-- & 92.0  \\
            GPT-4 (PHP) &40 &96.5  \\
            GPT-4 (Model~selection) &15 &96.8 \\
            \midrule
            \gptcode&-  & 92.9   \\
            {\shortstack{\gptcode}~+ \textbf{CSV} + \textbf{Voting}} & \textbf{5} & \textbf{97.0}  \\
        \bottomrule
	\end{tabular}}
    \vspace{-10pt}
	\label{tab:gsm8k}
    \end{minipage}
    \begin{minipage}[t]{0.01\textwidth}
    \end{minipage}
    \begin{minipage}[t]{0.49\textwidth}
    \centering
	\caption{Performances on MMLU dataset.}
    \resizebox{0.95\columnwidth}{!}{
	\begin{tabular}[b]{lccc }
		\toprule
		 Method & Dataset  & Accuracy(\%) & Few-shot \\
		\midrule
            Chinchilla~\citep{hoffmann2022training} &Math & 35.7 & 5-shot \\
            Galactica~\citep{taylor2022galactica} &Math & 41.3 &  5-shot   \\
            \midrule 
            \gptcode   &Math &87.5 & zero-shot \\
            {\shortstack{\gptcode}~+ \textbf{CSV} + \textbf{Voting}}  &Math & \textbf{89.2} & \textbf{zero-shot}  \\
            \midrule 
            LLaMA 2 &STEM &58.0 & 5-shot \\
            OpenLLM &STEM &70.6 & 5-shot \\
            GPT-4   &STEM &82.7 & zero-shot \\
            \gptcode   &STEM &86.8 & zero-shot \\
            {\shortstack{\gptcode}~+ \textbf{CSV} + \textbf{Voting}} &STEM  & \textbf{87.0} & \textbf{zero-shot} \\
        \bottomrule
	\end{tabular}}

    \label{tab:mmlu_math}
    \end{minipage}
\end{table}

\subsection{Code usage frequency of proposed prompts}

\begin{figure}[t]
    \centering
    \includegraphics[width=\textwidth]{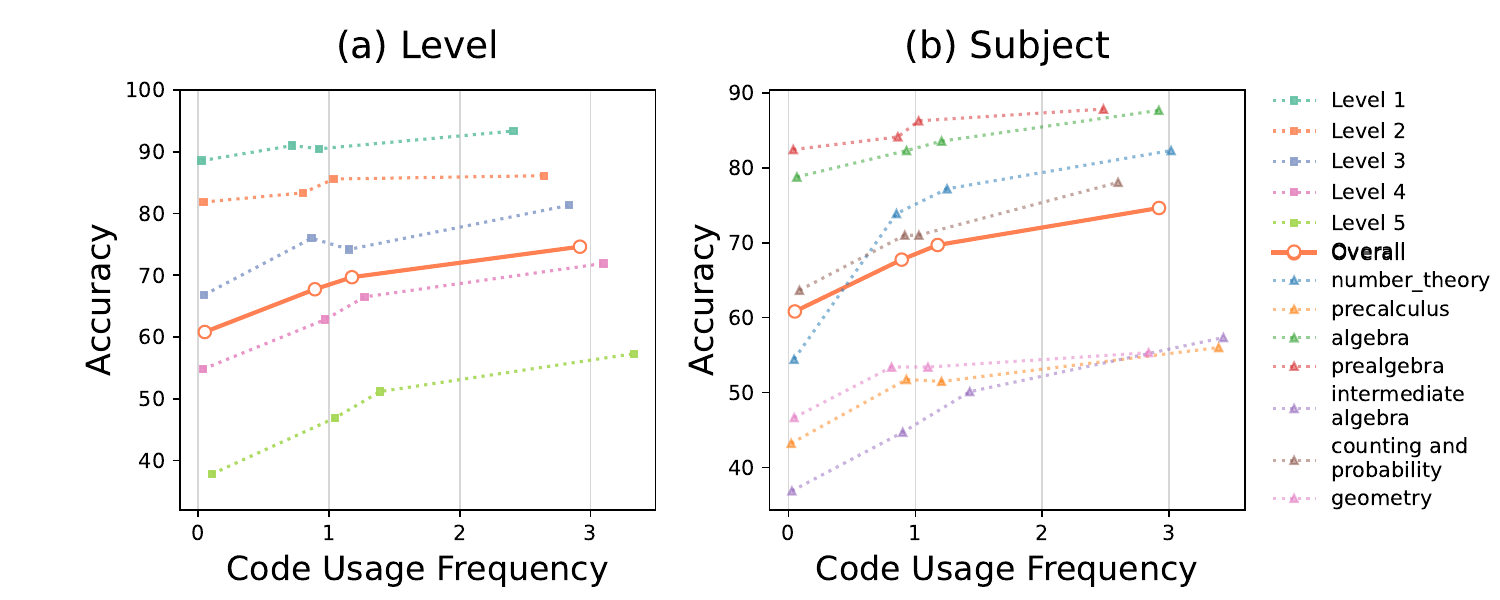}
    \vspace{-4mm}
    \caption{The four points on each curve correspond to results using \textbf{Prompt 1}, \textbf{Prompt 2}, \textbf{Basic Prompt} and \textbf{Code-based Self-verification Prompt}, respectively. \textbf{(a)} The accuracy of different levels at various code usage frequencies. \textbf{(b)} The accuracy of different subjects at various code usage frequencies.}
\label{fig:code-frequency}
\end{figure}

Analogous to the approach taken in Sec.~\ref{pilot-experiments}, we gather data to elucidate the correlation between accuracy and Code Usage Frequency across various dimensions - prompts (proposed CSV prompt as well as prompts used in pilot experiments), subjects, and difficulty levels. As shown in Fig.~\ref{fig:code-frequency}, 
the model's behavior is in accordance with our expectations when adding the code-based prompts. Each line in Fig.~\ref{fig:code-frequency} has an obvious trend of going upwards, proving that the increase of Code Usage Frequency induces a general improvement in accuracy. The performance gain when using more code is more obvious in the higher difficulty levels, while in lower levels, the performance gain is not very prominent, as shown in Fig.~\ref{fig:code-frequency} (a). Also, the Code Usage Frequency increases steadily with the increase of difficulty levels. This shows that the harder math problems require more frequent code usage, which implies that invoking code multiple times might be an important reason why \gptcode~have such an advantage in solving difficult math problems. There is a similar trend in Fig.~\ref{fig:code-frequency} (b).

\subsection{Ablation Study and Discussion}

\textbf{Comparisons between Natural Language and Code-based Self-Verification:} to underscore the significance of code in the self-verification stage, we employed a distinct natural language self-verification prompt. In this approach, \gptcode ~ is directed to verify the solution through natural language instead of relying on code-based verification, as presented in Tab.~\ref{tab:gpt4_math_verification}. The accuracy achieved with this method was slightly lower than that of the \textit{Basic Prompt}. Moreover, we observed a decline in accuracy for 4 of the 7 subtopics, indicating that relying solely on natural language verification can not only compromise accuracy but also negatively impact performance. In contrast, code-based verification enhances accuracy across all 7 subtopics when compared to the \textit{Basic Prompt}.

\begin{table}[t]
   \caption{Comparison Self-verification with/without explicit code-based prompt (\textbf{Overall}:The results across various MATH subtopics~\citep{hendrycks2021measuring})}
   \label{tab:gpt4_math_verification}
   \centering
   \resizebox{\columnwidth}{!}{
   \begin{tabular}{lccccccccc}
     \toprule
 & \textbf{Verification} & Intermediate & Precalculus &Geometry & Number &Counting \& &PreAlgebra  & Algebra &Overall  \\
 & \textbf{Method} &Algebra & \textbf{--}  & \textbf{--} & Theory & Probability & \textbf{--}  & \textbf{--}  & \textbf{--}  \\
     \cmidrule(r){2-2} \cmidrule(r){3-10}  
 
 \multirow{2}{*}{{\large \gptcode}}&without Verification  & 50.1&51.5&53.4&77.2&70.6&86.3&83.6&69.69 \\
 \cmidrule(r){2-10}
  \multirow{2}{*}{\large Interpreter}  &Nature Language &{52.6}&{48.7}&{50.8}&{79.9}&{72.5}&{83.1}&  {82.6}&{69.29}\\
 && \textcolor{c1}{\textbf{ +2.5}} & \textcolor{c2}{\textbf{ -2.8}} & \textcolor{c2}{\textbf{ -2.6}} & \textcolor{c1}{\textbf{ +2.7}} & \textcolor{c1}{\textbf{ +1.9}} & \textcolor{c2}{\textbf{ -3.2}} & \textcolor{c2}{\textbf{ -1.0}} & \textcolor{c2}{\textbf{ -0.40}} \\
 & Code-based  &\textbf{56.6}&\textbf{53.9}&\textbf{54.0}&\textbf{85.6}&\textbf{77.3}&\textbf{86.5}&\textbf{86.9}&\textbf{73.54} \\
 && \textcolor{c1}{\textbf{ +6.5}} & \textcolor{c1}{\textbf{+2.4}} & \textcolor{c1}{\textbf{ +0.6}} & \textcolor{c1}{\textbf{ +8.4}} & \textcolor{c1}{\textbf{ +6.7}} & \textcolor{c1}{\textbf{ +0.2}} & \textcolor{c1}{\textbf{ +3.3}} & \textcolor{c1}{\textbf{ +3.85}} \\

     \bottomrule
   \end{tabular}}
\end{table}

\begin{figure}[t]
    \centering
    \includegraphics[width=0.99\textwidth]{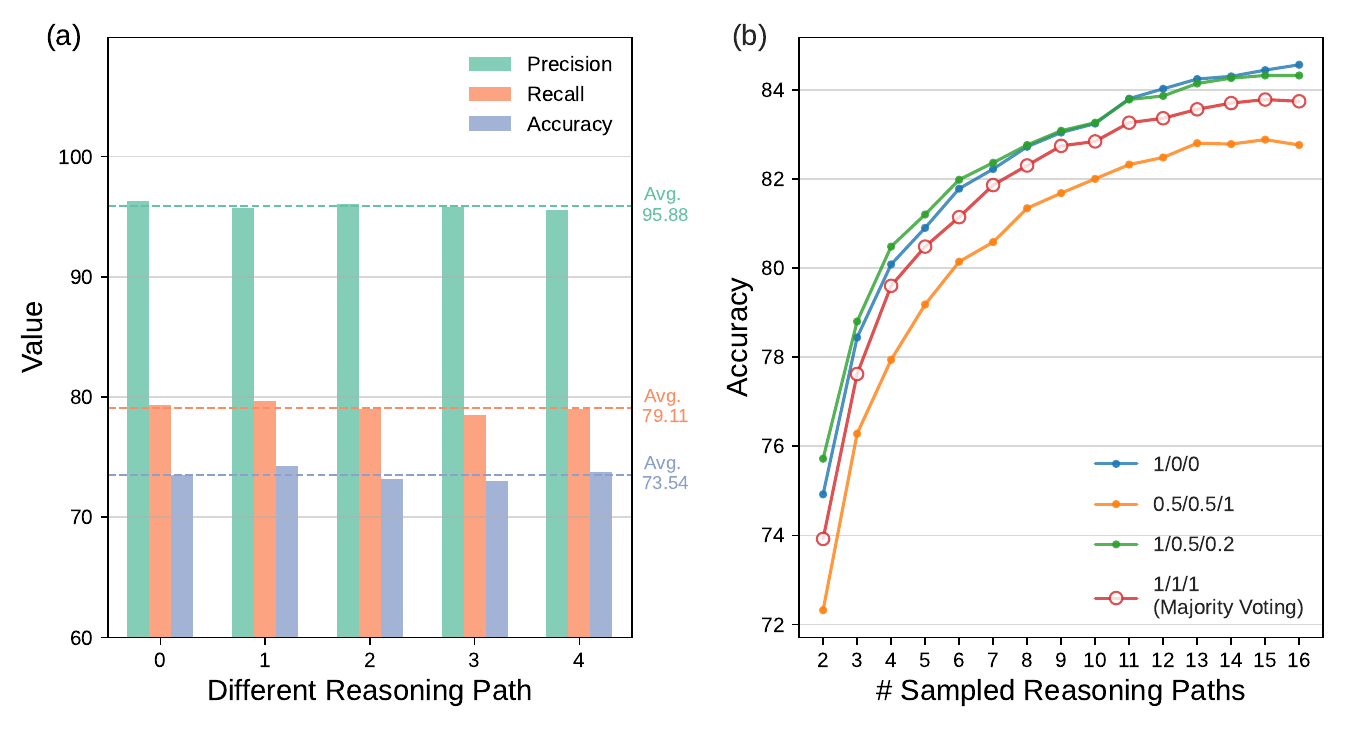}
    \vspace{-5mm}
    \caption{\textbf{(a).} shows the precision, recall, and accuracy on different reasoning paths. \textbf{(b).} shows the accuracy in response to the number of sampled reasoning paths when the weight is set to different values.}
\label{fig:analysis_voting}
\end{figure}

\textbf{Analysis of Verification-guided Weighted Majority Voting:} we initially compiled the confusion matrix (TP/TN/FP/FN), capturing solutions with self-verification that matches the \textit{True} and \textit{False} states mentioned in Eq.~\ref{eq1} from five distinct sampled paths. The details of the confusion matrix are presented in Appendix~\ref{sec:conf}. From this data, we computed Precision, Recall, and Accuracy. (Solutions in the \textit{True} state are seen as positive.) The results are presented in Fig.~\ref{fig:analysis_voting}. In comparison to Accuracy, we observed numerical enhancements of 22.3\% and 5.6\% in the average Precision and Recall, respectively. In particular, the average Precision registered at 95.88\%. This implies that the Accuracy has the potential to become much higher, if more solutions reach the verified \textit{True} state before giving the final answer. 

\textbf{Hyper-parameters ablation in Verification-guided Weighted Majority Voting:} we also performed ablation studies on the hyper-parameter $w_v \in \{w_\mathbf{\scriptscriptstyle{T}}, w_\mathbf{{\scriptscriptstyle{U}}}, w_\mathbf{{\scriptscriptstyle{F}}}\}$ in Eq.~\ref{eq1}. When the hyper-parameter setting satisfied ${w_\mathbf{\scriptscriptstyle{T}} > w_\mathbf{{\scriptscriptstyle{U}}} \geq w_\mathbf{{\scriptscriptstyle{F}}}}$, the performance of the verification-guided weighted majority voting consistently surpassed that of the naive majority voting methods across all sampled paths. In contrast, when we set the hyper-parameter $(w_\mathbf{\scriptscriptstyle{T}}=0.5, w_\mathbf{{\scriptscriptstyle{U}}}=0.5, w_\mathbf{{\scriptscriptstyle{F}}}=1)$, the performance under this configuration was worse than the naive majority voting. Therefore, our proposed method, verification-guided weighted majority voting, is easy to tune and robust.

\section{Conclusion and Limitation}

In this paper, we begin with pilot experiments on \gptcode~to explore how its use of code impacts its performance in mathematical reasoning. By analyzing Code Usage Frequency and accuracy, we determine that \gptcode's skill in solving math problems can be largely attributed to its ability to generate and execute code, as well as its effectiveness in adjusting and rectifying solutions when confronted with implausible execution outputs. Expanding on this understanding, we introduce the ideas of explicit code-based self-verification and verification-guided weighted majority voting, with the goal of enhancing \gptcode's mathematical capabilities.

However, there are limitations in our work that we plan to explore further in the future. Firstly, our analysis and improvements are currently focused on \gptcode, which is somewhat restrictive. We aim to apply the methods to other LLMs. Secondly, our explicit code-based self-verification and verification-guided weighted majority voting technique could potentially create more accurate datasets. These datasets would include detailed step-by-step code-based solution generation and code-based validation, which could help improve open-source LLMs like LLaMA 2~\citep{touvron2023llama} and enhance their mathematical abilities. Although we haven't yet investigated this approach, we leave it for future work.

\newpage
\bibliography{iclr2023_conference}
\bibliographystyle{iclr2023_conference}
\newpage
\appendix

\section*{Appendix}

This Appendix contains two sections. The first section provides the experiment details, including the detailed experiment results on MATH and MMLU datasets. The second section presents some examples of \gptcode.

\section{Experiment Details}

\subsection{Detailed experiment result on MATH dataset}
\subsubsection{Confusion Matrix}
\label{sec:conf}
A confusion matrix is a specific table layout that allows visualization of the performance of an algorithm. It's particularly useful for classification problems, and we utilize it to analyze the performance of our verification process.

The matrix itself is a two-dimensional grid, 2x2, for the binary classification of verification results. Each row of the matrix represents the instances in a predicted class, which is determined by the verification results given by the language model, while each column represents the instances in an actual class, which is determined by the actual correctness of the answer given by the model. Tab.~\ref{tab:confision_matrix} shows how the matrix looks for our verification process:

\begin{table}[ht]
\begin{center}
\caption{Confusion Matrix of Verification}
\label{tab:confision_matrix}
\begin{tabular}{c|cc}

\toprule
	& Answer Correct	& Answer Wrong\\
\midrule
Verification True	& TP & FP \\
Verification False & FN & TN \\
\bottomrule

\end{tabular}
\end{center}
\end{table}
Here's what the four terms mean:
\begin{itemize}
    \item True Positive (TP): The cases in which the model's verification result is ‘True', and the answer is actually correct.
    \item True Negative (TN): The cases in which the model's verification result is ‘False', and the answer is actually wrong.
    \item False Positive (FP): The cases in which the model's verification result is ‘True', but the answer is actually wrong.
    \item False Negative (FN): The cases in which the model's verification result is ‘False', but the answer is actually correct.
\end{itemize}

This matrix is very helpful for measuring more than just straightforward accuracy, based on which Precision and Recall are two important metrics. 
They are defined in Eq.~\ref{eq:pr} and their meanings are as follows: 
\begin{itemize}
\item Precision is the fraction of relevant instances among the retrieved instances. It is a measure of the accuracy of the classifier when it predicts the positive class.
\item Recall is the fraction of the total amount of relevant instances that were actually retrieved. It is a measure of the ability of a classifier to find all the positive instances.
\end{itemize}

\begin{align}
\label{eq:pr}
    \text{Precision}=\frac{\text{TP}}{\text{TP}+\text{FP}}, \;    \text{Recall}=\frac{\text{TP}}{\text{TP}+\text{FN}}
\end{align}

In other words, precision answers the question “What proportion of verified TRUE answers was actually correct?", while recall answers “What proportion of actual correct answers was verified TRUE?"
Given its meaning, verification-guided voting is bounded to be effective when the precision of verification is high.

\subsubsection{Python package usage analysis}

Tab.~\ref{tab:pythonpack} outlines the usage of various Python packages in our experiments. Among them, we found that the \texttt{sympy} package is utilized most frequently, highlighting its central role in the computational tasks performed.

\begin{table}[ht]
\caption{Python package usage frequency on MATH dataset.}
\label{tab:pythonpack}
\begin{center}
\resizebox{\columnwidth}{!}{
\begin{tabular}{cccccccccccc}
\\ \toprule 
Package             & sympy  & numpy  & math   & fractions & itertools & cmath  & scipy  & matplotlib & functools & collections & statistics \\ \midrule
All            & 0.4168 & 0.0284 & 0.1590  & 0.0094 & 0.0034 & 0.0034 & 0.0016 & 0.0010  & 0.0004 & 0.0004 & 0.0002 \\
Correct        & 0.3907 & 0.0241 & 0.1638 & 0.0110  & 0.0029 & 0.0026 & 0.0009 & 0.0003 & 0.0003 & 0.0006 & 0.0003 \\
correct per code & 0.3323 & 0.0205 & 0.1393 & 0.0094    & 0.0025    & 0.0022 & 0.0007 & 0.0003     & 0.0003    & 0.0005      & 0.0003     \\
Wrong          & 0.4724 & 0.0383 & 0.1493 & 0.0058 & 0.0045 & 0.0052 & 0.0032 & 0.0026 & 0.0007 & 0      & 0      \\
wrong per code & 0.3194 & 0.0259 & 0.1009 & 0.004  & 0.0031 & 0.0035 & 0.0022 & 0.0018 & 0.0004 & 0      & 0      \\
c/w per code   & 104\%  & 79\%   & 138\%  & 238\%  & 80\%   & 63\%   & 34\%   & 14\%   & 57\%   & NaN    & NaN   \\
\bottomrule
\end{tabular}}
\end{center}
\end{table}

\subsection{Detailed experiment result on MMLU dataset}

Fig.~\ref{fig:mmlu} illustrates that \gptcode~performs relatively poorly in certain domains, such as engineering and the humanities, with a particularly marked deficiency in virology, where it achieves a score of less than 60\%. These observations delineate specific areas that call for further investigation and refinement, thus outlining the direction for future improvements in the model.

\begin{figure}[t]
    \centering
    \includegraphics[width=0.9\textwidth]{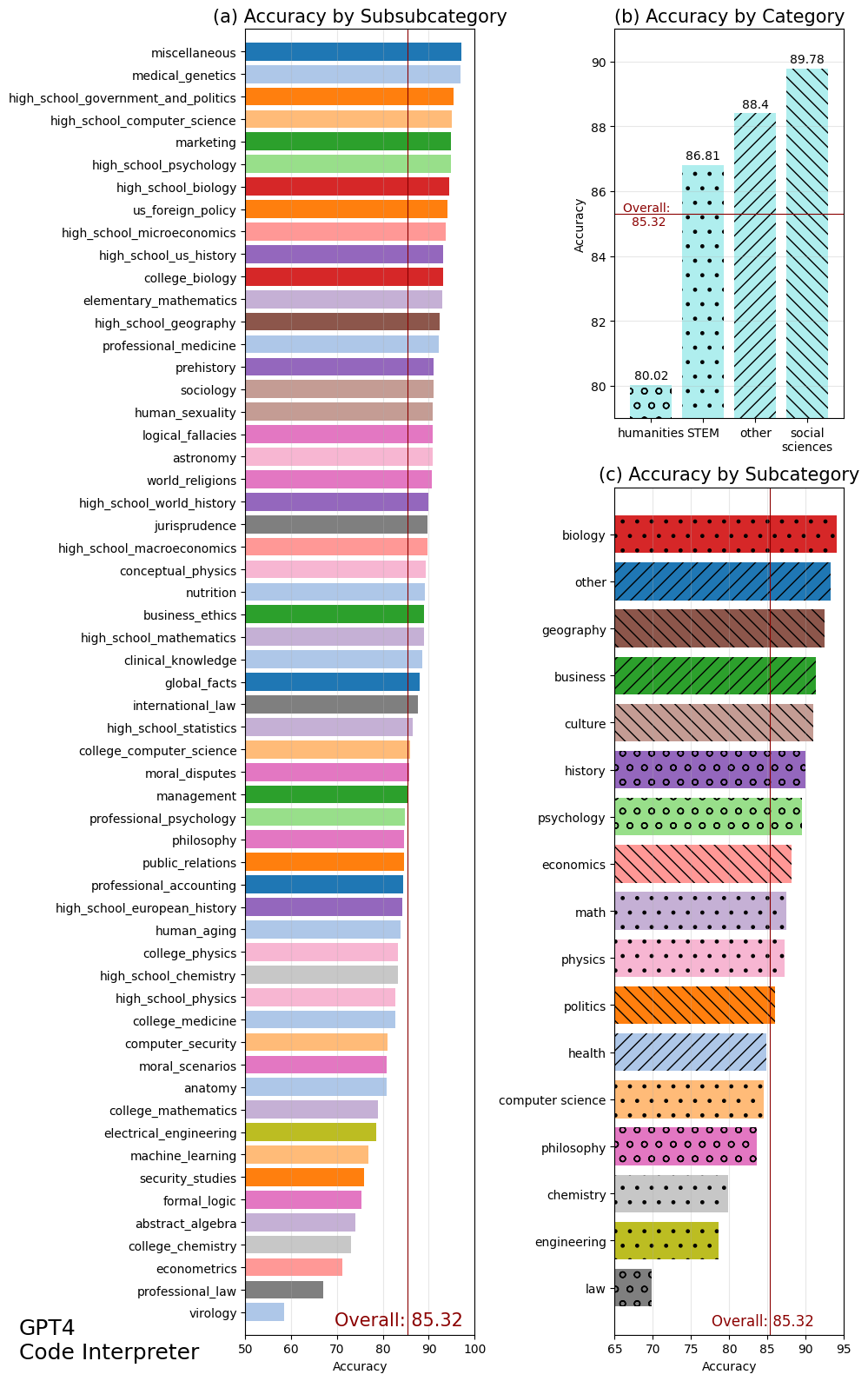}
    \caption{\gptcode~performance on MMLU dataset on: (a) subsubcategory, (b) category, (c) subcategory}
\label{fig:mmlu}
\end{figure}

\begin{figure}[t]
    \centering
    \includegraphics[width=0.9\textwidth]{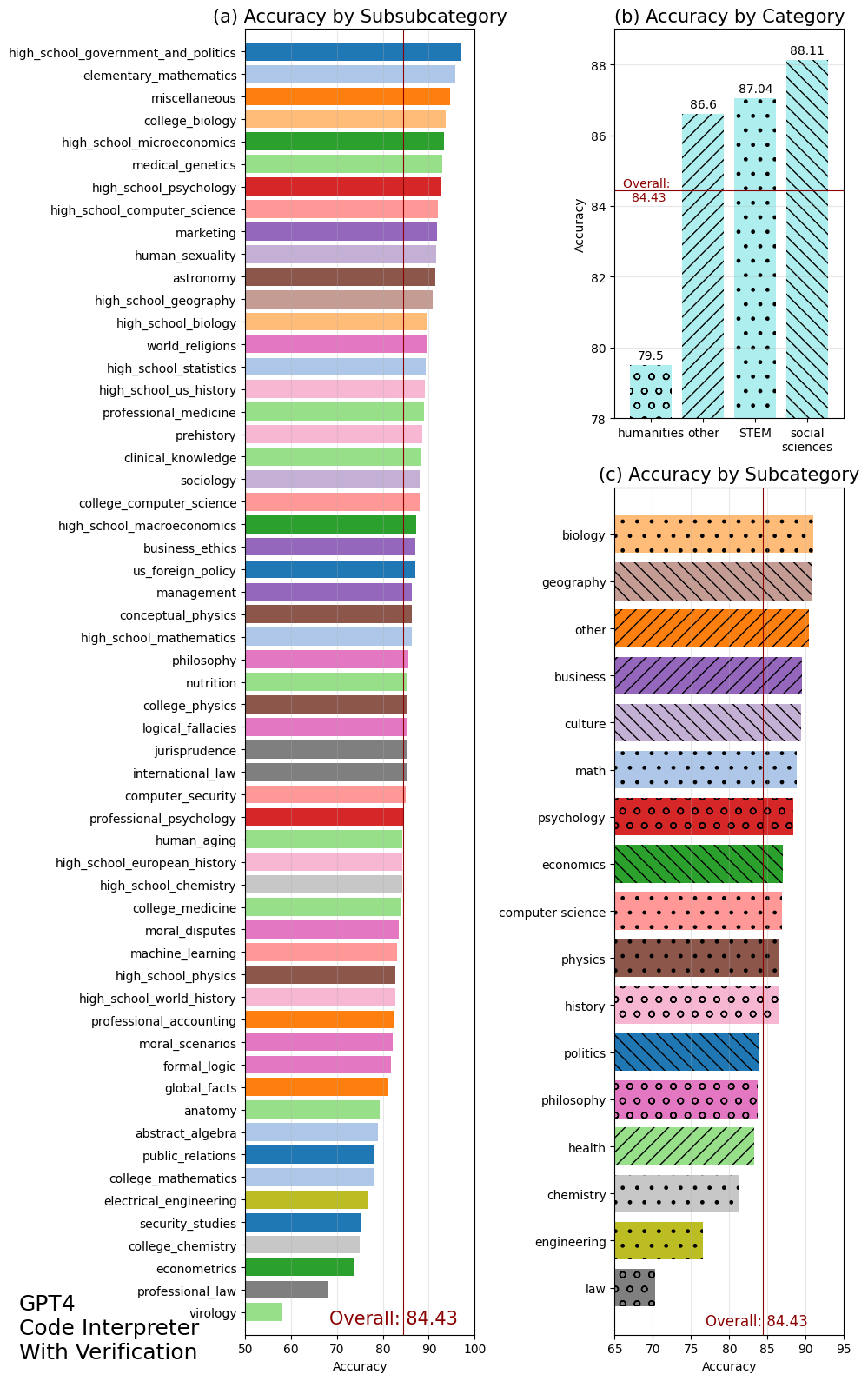}
    \caption{Model performance on MMLU dataset on: (a) subsubcategory, (b) category, (c) subcategory}
\label{fig:mmlu}
\end{figure}

\section{Examples}

In this section, we provide more examples.

\begin{table*}[ht]
    \centering
    \begin{tabular}{p{0.95\textwidth}}
    \toprule
     \multicolumn{1}{p{0.95\textwidth}}{\small \textbf{Math Problem}:  $0.\overline{54}$ as a fraction in lowest terms.} \\
     
    \midrule
Sure, we can solve this problem by setting up an equation, solving it, and simplifying the fraction.

Let's call \(0.\overline{54}\) as \(x\).

The decimal \(0.\overline{54}\) means that the digits "54" repeat indefinitely. So, we can multiply \(x\) by 100 (since "54" is a two-digit number) to get another equation.

This gives us \(100x = 54.\overline{54}\).

By subtracting the original equation from this new one, we eliminate the repeating decimal:

\(100x - x = 54.\overline{54} - 0.\overline{54} \).

Let's perform these steps in the next Python cell.

\lstset{style=mystyle}
     \begin{lstlisting}[language=Python]
# define the symbol
x = symbols('x')

# define the equations
eq1 = Eq(x, 0.54545454545454545454545454545454)  # approximating the repeating decimal
......
     \end{lstlisting} 
     \textbf{RESULT}\newline
          \begin{lstlisting}[language=Python]
---------------------------------------------------------------------
TypeError                           Traceback (most recent call last)
......
          \end{lstlisting} 
Apologies for the mistake. We can't directly subtract two equations in Python. Instead, we should subtract the right-hand sides and left-hand sides of the equations separately to get a new equation. Let's correct that.
\lstset{style=mystyle}
     \begin{lstlisting}[language=Python]
# subtract the right hand sides and left hand sides of the equations separately
rhs = simplify(eq2.rhs - eq1.rhs)
lhs = simplify(eq2.lhs - eq1.lhs)
......
     \end{lstlisting} 
     \textbf{RESULT}\newline
    \textbf{Eq(99*x, 54.0)}\newline
    After subtraction, we get the equation \(99x = 54\). 

Next, we can solve for \(x\) to express \(0.\overline{54}\) as a fraction.
\lstset{style=mystyle}
     \begin{lstlisting}[language=Python]
# solve for x
solution = solve(new_eq, x)

solution[0]
     \end{lstlisting} 
     \textbf{RESULT}\newline
    \textbf{0.545454545454545}\newline
The solution for \(x\) is approximately \(0.545454545454545\), but we need to express it as a fraction. Let's convert this decimal into a fraction.
\lstset{style=mystyle}
     \begin{lstlisting}[language=Python]
# convert the decimal solution to a fraction
fraction = nsimplify(solution[0])

fraction
     \end{lstlisting} 
     \textbf{RESULT}\newline
    \textbf{6/11}\newline
The decimal \(0.\overline{54}\) can be expressed as the fraction \(\frac{6}{11}\).

     \\
     \midrule

      \multicolumn{1}{p{0.90\textwidth}}{\small \textbf{Answer}:  \(\frac{6}{11}\)}
\\
     \bottomrule
    \end{tabular}
    \caption{An example in which \gptcode~self-debugs when the code execution emits an error.}
    \label{tab:baseline_example1}
\end{table*}

\begin{table*}[ht]
    \centering
    \begin{tabular}{p{0.95\textwidth}}
    \toprule
     \multicolumn{1}{p{0.95\textwidth}}{\small \textbf{Math Problem}:  A bus comes by Jerry's bus stop every 20 minutes starting at exactly 5:13 a.m. If Jerry shows up at exactly 8:35 a.m., how many minutes will he have to wait for the next bus?} \\
     
    \midrule
Step 1: Calculate the total minutes from 5:13 a.m. to 8:35 a.m.

Total minutes from 5:13 a.m. to 8:35 a.m. is \(180 + 22\) minutes.

Step 2: Calculate how many buses have come by during this time.

Step 3: Calculate the time when the last bus before 8:35 a.m. came by.

- The time when the last bus came by is \(5:13 \, \text{{a.m.}} + 20 \times \lfloor n \rfloor \, \text{{minutes}}\).

Step 4: Calculate the waiting time.

Now, let's calculate each step.
\lstset{style=mystyle}
     \begin{lstlisting}[language=Python]
# Step 1: Calculate the total minutes from 5:13 a.m. to 8:35 a.m.
total_minutes = 180 + 22
......
     \end{lstlisting} 
     \textbf{RESULT}\newline
    \textbf{-311}\newline
It seems that there's a mistake in the calculations. The waiting time cannot be negative. This error might have occurred in Step 3, where I calculated the time of the last bus in minutes past midnight, but then in Step 4, I subtracted this from the total time from 5:13 a.m. to 8:35 a.m. 

I should correct Step 4 to calculate the time until the next bus, which is either 0 if a bus has just arrived at 8:35 a.m., or 20 minutes minus the time since the last bus if a bus arrived before 8:35 a.m.

Let's correct this.
\lstset{style=mystyle}
     \begin{lstlisting}[language=Python]
# Corrected Step 4: Calculate the time until the next bus
time_since_last_bus = total_minutes - last_bus_time
......
     \end{lstlisting} 
     \textbf{RESULT}\newline
    \textbf{331}\newline
There still seems to be a mistake in the calculations. The waiting time calculated is significantly more than 20 minutes, which is not possible given that the bus comes every 20 minutes. 

Looking back at the calculations, it seems that the error is in Step 3 when calculating the time the last bus came. The last bus time should be calculated from 5:13 a.m., not from midnight. 

Let's correct this.
\lstset{style=mystyle}
     \begin{lstlisting}[language=Python]
# Corrected Step 3: Calculate the time when the last bus before 8:35 a.m. came by.
last_bus_time_corrected = 5 * 60 + 13 + num_buses * 20
......
     \end{lstlisting} 
     \textbf{RESULT}\newline
    \textbf{331}\newline
It seems there is still an error. Let's try a different approach. 

1. Calculate the total minutes that have passed since the first bus at 5:13 a.m. until 8:35 a.m.

2. Since the buses come every 20 minutes, calculate how many complete cycles of 20 minutes have passed in the calculated total minutes.

3. The time Jerry will have to wait is the remainder of the total minutes divided by 20 (which is the time since the last bus) subtracted from 20 (which is the bus cycle).

This approach should give us the correct waiting time. Let's calculate it.
\lstset{style=mystyle}
     \begin{lstlisting}[language=Python]
# Calculate total minutes from 5:13 a.m. to 8:35 a.m.
total_minutes = 3 * 60 + 22  # 3 hours and 22 minutes
......
     \end{lstlisting} 
     \textbf{RESULT}\newline
    \textbf{18}\newline
    So, Jerry will have to wait for 18 minutes for the next bus. 

     \\
     \midrule

      \multicolumn{1}{p{0.90\textwidth}}{\small \textbf{Answer}:  18}
\\
     \bottomrule
    \end{tabular}
    \caption{An example in which \gptcode~adjusts its solution based on the result of code execution.}
    \label{tab:baseline_example2}
\end{table*}

\begin{table*}[ht]
    \centering
    \begin{tabular}{p{0.30\textwidth}|p{0.30\textwidth}|p{0.30\textwidth}}
    \toprule
     \multicolumn{3}{p{0.90\textwidth}}{\small \textbf{Math Problem}: Solve for $x: 3^{2x} + 19 = 10^x$.} \\
     
    \midrule

    \small \textbf{code 0} 
\newline\newline
    \lstset{style=mystyle}
    \small
     This is a difficult equation to solve and there's no simple analytical solution using elementary functions. In real world scenarios, such equations are typically solved numerically. For example, using a graphing calculator, a computer, or a mathematical software.

However, if you want an approximation of the solution, it is advisable to use numerical methods such as the bisection method, Newton's method, or others.

If you would like me to provide an algorithm to find the solution numerically, let me know.
\small
     &
    \small \textbf{code 1} \newline
    \newline
    \small Answer: \newline
     \small  
The equation you're asking to solve, $3^{2x} + 19 = 10^x$, cannot be solved algebraically due to the nature of the functions and operations involved. Therefore, we'll need to use numerical methods such as Newton's method or Bisection method for root finding.\newline

\lstset{style=mystyle}
     \begin{lstlisting}[language=Python]
from scipy.optimize import fsolve
import numpy as np

# Defining the function
def func(x):
    return 3**(2*x) + 19 - 10**x

# Initial guess for x, it can be anything, 1.0 is chosen arbitrarily here
x0 = 1.0

# Using fsolve to find the root
root = fsolve(func, x0)

x = root[0]
x
     \end{lstlisting} 
     \textbf{RESULT}\newline
     \textbf{1.001831315399663}\newline
     So the answer is 1.00.
     &
     \small \textbf{code more} \newline
     \newline
     \small Answer: \newline
     \small 
    Rewrite the equation as:\newline
    $3^{2x}-10+19=0$\newline
    Let's denote $3^x=a$, and $10^x=b$\newline
    so $a^2-b+19=0$\newline
    now we have two equations:\newline
    1. $a=3^x$\newline
    2. $b=10^x$
    Let's solve this with Python.
    \lstset{style=mystyle}
     \begin{lstlisting}[language=Python]
from scipy.optimize import root
import numpy as np

# Define the system of equations
def equations(vars):
    x, a, b = vars
    eq1 = a - 3**x
    eq2 = b - 10**x
    eq3 = a**2 - b + 19
    return [eq1, eq2, eq3]
.....
     \end{lstlisting} 
    \textbf{RESULT:}\newline
    \textbf{0.0}\newline
    $3^{2.0}+19=10^0$\newline
The left-hand side becomes 1+19=20 and the right-hand side becomes 1. Thus, the original equation does not hold true for x=0.\newline
     Let's try a different initial guess for the solution to see if that helps.
         \lstset{style=mystyle}
     \begin{lstlisting}[language=Python]
# Adjust initial guess for x, a, b
x0 = [1, 3, 10]

# Solve system of equations
solution = root(equations, x0)

# Extract solution
x = solution.x[0]
x
     \end{lstlisting} 
         \textbf{RESULT:}\newline
    \textbf{2.0000000000004463}\newline
    We know that $3^{2\cdot2}=3^4=81$ and $10^2=100$. The left-hand side becomes 100 and the right-hand side is also 100. Thus, the original equation holds true for $x=2$. 
     \\
     \midrule

      \multicolumn{3}{p{0.90\textwidth}}{\small \textbf{Answer}: 2}
\\
     \bottomrule
    \end{tabular}
    \caption{An example in which no restriction on code use succeeds while not using code and using code only once fails.}
    \label{tab:code_freq_example1}
\end{table*}

\begin{table*}[ht]
    \centering
    \begin{tabular}{p{0.30\textwidth}|p{0.30\textwidth}|p{0.30\textwidth}}
    \toprule
     \multicolumn{3}{p{0.90\textwidth}}{\small \textbf{Math Problem}: What is the sum of the digits in the terminating decimal representation of the fraction $\frac{4321}{5^7\cdot2^8}$?} \\
     
    \midrule

    \small \textbf{code 0} 
\newline
\newline
    \lstset{style=mystyle}
    \small
     The given fraction is 

\[
\frac{4321}{5^7\cdot2^8}
\]

To simplify this fraction, we can notice that \(5^7\cdot2^8 = 10^7\), which would mean the fraction is equivalent to \(4321 \times 10^{-7}\), i.e., the number 4321 with the decimal point moved 7 places to the left. This would result in the number 0.0004321.

To get the sum of the digits in the terminating decimal representation of this number, we just need to add up the digits in 4321, since the zeros do not contribute to the sum.

So, the sum is \(4 + 3 + 2 + 1 = 10\).

Therefore, the answer is 10\newline
\small
     &
    \small \textbf{code 1} \newline
    \newline
    \small Answer: \newline
     \small  
First, we need to simplify the fraction. 

Given that \(4321 = 2^5 \cdot 3^3 \cdot 5^0 \cdot 7^0 \cdot 11^0 \cdot 13^1\), we can simplify the fraction \(\frac{4321}{5^7\cdot2^8}\) to \(\frac{2^5 \cdot 3^3 \cdot 5^0 \cdot 7^0 \cdot 11^0 \cdot 13^1}{5^7\cdot2^8}\) = \(3^3 \cdot 13^1 \cdot 2^{-3} \cdot 5^{-7}\).\newline
Next, we need to compute the decimal representation and sum the digits.
    \lstset{style=mystyle}
     \begin{lstlisting}[language=Python]
# Compute the decimal representation of the fraction
value = (3**3 * 13**1) / (2**3 * 5**7)

# Convert to string and replace the decimal point
str_value = str(value).replace('.', '')

# Compute the sum of the digits
digit_sum = sum(int(digit) for digit in str_value)

digit_sum
     \end{lstlisting} 
    \textbf{RESULT:}\newline
    \textbf{18}\newline
     &
     \small \textbf{code more} \newline
     \newline
     \small Answer: \newline
     \small 
    The denominator can be rewritten as \(10^{min(7,8)} \times 5^{max(0,7-8)} \times 2^{max(0,8-7)}\), where \(min(a,b)\) is the minimum of \(a\) and \(b\), and \(max(a,b)\) is the maximum of \(a\) and \(b\). Now let's calculate the value of \(min(7,8)\), \(max(0,7-8)\), and \(max(0,8-7)\).
        \lstset{style=mystyle}
     \begin{lstlisting}[language=Python]
min_ab = min(a, b)
max_0_a_minus_b = max(0, a-b)
max_0_b_minus_a = max(0, b-a)

min_ab, max_0_a_minus_b, max_0_b_minus_a
     \end{lstlisting} 
    \textbf{RESULT:}\newline
    \textbf{(7, 0, 1)}\newline
    Now, let's simplify the fraction \( \frac{4321}{10^7 \times 5^0 \times 2^1} \).
         \begin{lstlisting}[language=Python]
# define the numerator and denominator
numerator = 4321
denominator = 10**7 * 5**0 * 2**1
......
     \end{lstlisting} 
    \textbf{RESULT:}\newline
    \textbf{0.00021605}\newline

We exclude the zeros before the first non-zero digit and the decimal point as they do not contribute to the sum.

Thus, the sum of the digits is the sum of \(2\), \(1\), \(6\), \(0\), and \(5\). Let's calculate it.
        \lstset{style=mystyle}
     \begin{lstlisting}[language=Python]
# digits in the fraction
digits = [2, 1, 6, 0, 5]

# calculate the sum of the digits
sum_of_digits = sum(digits)
sum_of_digits
     \end{lstlisting} 
    \textbf{RESULT:}\newline
    \textbf{14}\newline
    So, the final answer is: 14
     \\
     \midrule

      \multicolumn{3}{p{0.90\textwidth}}{\small \textbf{Answer}: 14}
\\
     \bottomrule
    \end{tabular}
    \caption{An example in which no restriction on code use succeeds while not using code and using code only once fails.}
    \label{tab:code_freq_example2}
\end{table*}

\begin{table*}[ht]
    \centering
    \begin{tabular}{p{0.30\textwidth}|p{0.30\textwidth}|p{0.30\textwidth}}
    \toprule
     \multicolumn{3}{p{0.90\textwidth}}{\small \textbf{Math Problem}:  What is the sum of all positive integers $r$ that satisfy $\mathop{\text{lcm}}[r,700] = 7000~?$} \\
     
    \midrule

    \small \textbf{True} 
\newline\newline
The least common multiple (LCM) of two numbers \( a \) and \( b \) is given by the formula:
\[
\text{LCM}(a, b) = \frac{|a \times b|}{\text{GCD}(a, b)}
\]

Hence, the equation can be rewritten as:
\[
\frac{r \times 700}{\text{GCD}(r, 700)} = 7000
\]

We also know that \( r \) must be a divisor of 7000. So, we'll find all the divisors of 7000.\newline
\lstset{style=mystyle}
     \begin{lstlisting}[language=Python]
# Find all divisors of 7000
divisors_7000 = sympy.divisors(7000)
# Filter out the divisors
r_values_corrected = [r for r in divisors_7000 if r == 7000 * sympy.gcd(r, 700) // 700]
......
     \end{lstlisting} 
    \textbf{RESULT:}\newline
    \textbf{(7000, 1000)}\newline
    Let's verify our solution by checking whether the LCM of each possible \( r \) and 700 is indeed 7000.
    \lstset{style=mystyle}
     \begin{lstlisting}[language=Python]
# Verify the solution
lcm_values_corrected = [sympy.lcm(r, 700) for r in r_values_corrected]
......
     \end{lstlisting} 
    \textbf{RESULT:}\newline
    \textbf{(True, True)}\newline
    Upon verification, we find that the least common multiple (LCM) of each possible value of \( r \) and 700 is indeed 7000.
    Hence, the sum of all positive integers \( r \) that satisfy \(\text{lcm}[r, 700] = 7000\) is \(1000+7000=8000\).
    The answer is 8000.
\small

     &
     \small \textbf{Uncertain} \newline
     \newline
          First, let's find the prime factorization of 7000. \newline
     Then, we'll generate all the factors of 7000 and find which of those factors are multiples of 700. \newline
     Finally, we'll sum up all the positive integers that satisfy the condition. \newline
     \lstset{style=mystyle}
     \begin{lstlisting}[language=Python]
import sympy
# Get the prime factorization of 7000
prime_factors = sympy.factorint(7000)
# Generate all factors of 7000
factors = set()
for factor_2_count in range(prime_factors[2] + 1):
    for factor_5_count in range(prime_factors[5] + 1):
......
     \end{lstlisting} 
    \textbf{RESULT:}\newline
    \textbf{({2: 3, 5: 3, 7: 1}, [3500, 700, 7000, 1400], 12600)}\newline
    The prime factorization of 7000 is \(2^3 \cdot 5^3 \cdot 7^1\). 
The factors of 7000 that are multiples of 700 are: 3500, 700, 7000, and 1400.
So, the sum of all positive integers \(r\) that satisfy \(\text{lcm}(r, 700) = 7000\) is \(3500 + 700 + 7000 + 1400 = 12600\).\newline
So the final answer is 12600.
\newline
    &
        \small \textbf{False} 
\newline\newline
        \lstset{style=mystyle}
    \small
     In this problem, we are given that $lcm(r,700)=7000$.\newline
     Let's first find the prime factors of 700 and 7000.
          \lstset{style=mystyle}
     \begin{lstlisting}[language=Python]
import sympy

# Find the prime factorization of 700 and 7000
factors_700 = sympy.factorint(700)
......
     \end{lstlisting} 
    \textbf{RESULT:}\newline
    \textbf{({2: 2, 5: 2, 7: 1}, {2: 3, 5: 3, 7: 1})}\newline
    Since the LCM of r and 700 is 7000, r must contain all the prime factors of 7000 that are not already in 700.
          \lstset{style=mystyle}
     \begin{lstlisting}[language=Python]
# Generate all combinations
combinations = list(product(*[[1, i] for i in extra_factors]))
......
     \end{lstlisting} 
    \textbf{RESULT:}\newline
    \textbf{([1, 5, 2, 10], 18)}\newline
    The possible values of r are 1, 2, 5, and 10. \newline
    To verify our answer, we can check that the least common multiple of r and 700 is indeed 7000.
    \lstset{style=mystyle}
     \begin{lstlisting}[language=Python]
# Remove 1 from the possible values of r
possible_r_values.remove(1)
......
     \end{lstlisting} 
    \textbf{RESULT:}\newline
    \textbf{False}\newline
    It seems that there's a mistake. Not all the calculated values of r satisfy the given condition. \newline
    Let's correct the calculation.
    \lstset{style=mystyle}
     \begin{lstlisting}[language=Python]
# All the values of r should have the factors of 700
r_values = [700]
......
     \end{lstlisting} 
    \textbf{RESULT:}\newline
    \textbf{([700, 1400, 3500, 7000], 18720)}\newline
    Therefore, answer is 18720.
     \\
     \midrule

      \multicolumn{3}{p{0.90\textwidth}}{\small \textbf{Answer}: 8000}
\\
     \bottomrule
    \end{tabular}
    \caption{An example in which \textit{True} succeeds while \textit{Uncertain} and \textit{False} fails.}
    \label{tab:verify_example1}
\end{table*}

\begin{table*}[ht]
    \centering
    \begin{tabular}{p{0.30\textwidth}|p{0.30\textwidth}|p{0.30\textwidth}}
    \toprule
     \multicolumn{3}{p{0.90\textwidth}}{\small \textbf{Math Problem}:  Let $A = (1,-11,2),$ $B = (3,-4,1),$ and $C = (-2,1,-1).$  Compute $\angle ABC,$ in degrees. } \\
     
    \midrule

    \small \textbf{True} 
\newline\newline
The angle \( ABC \) is the angle between the vectors \( \vec{BA} \) and \( \vec{BC} \). We can find these vectors as follows:

\[
\vec{BA} = \vec{A} - \vec{B}= (-2, -7, 1)
\]

\[
\vec{BC} = \vec{C} - \vec{B} = (-5, 5, -2)
\]

The angle \( \theta \) between two vectors \( \vec{u} \) and \( \vec{v} \) can be found using the formula

\[
\cos(\theta) = \frac{\vec{u} \cdot \vec{v}}{||\vec{u}|| ||\vec{v}||}
\]
Let's calculate these values.
    \lstset{style=mystyle}
     \begin{lstlisting}[language=Python]
import numpy as np

# Define vectors
vec_BA = np.array([1, -11, 2]) - np.array([3, -4, 1])
......
     \end{lstlisting} 
    \textbf{RESULT:}\newline
    \textbf{120.00000000000001}\newline
    Let's now verify this answer.
    \[
\vec{u} \cdot \vec{v} = ||\vec{u}|| \cdot ||\vec{v}|| \cdot \cos(\theta)
\]

Substituting our values into this formula, we should get the same dot product (-27) that we calculated earlier. Let's check this.
    \lstset{style=mystyle}
     \begin{lstlisting}[language=Python]
# Compute the dot product using the formula
dot_product_verification = magnitude_BA * magnitude_BC * np.cos(theta * np.pi / 180)
......
     \end{lstlisting} 
    \textbf{RESULT:}\newline
    \textbf{-27}\newline
    which matches the dot product we calculated earlier. Therefore, the computed angle of \(120^\circ\) is correct.

\small

     &
     \small \textbf{Uncertain} \newline
     \newline
the cosine of the angle \(\theta\) between them is given by

\[
\cos \theta = \frac{\mathbf{u} \cdot \mathbf{v}}{||\mathbf{u}|| \, ||\mathbf{v}||}
\]
In the given problem, we want to find the angle between vectors \(\overrightarrow{AB}\) and \(\overrightarrow{BC}\), where

\[
\overrightarrow{AB} = \mathbf{B} - \mathbf{A}  = (2,7,-1)
\]
\[
\overrightarrow{BC} = \mathbf{C} - \mathbf{B} = (-5,5,-2)
\]
Now, we'll calculate the angle using the formula for \(\cos \theta\).
        \lstset{style=mystyle}
     \begin{lstlisting}[language=Python]
from numpy import array, dot
from numpy.linalg import norm
import math

# Defining the vectors
A = array([1, -11, 2])
......
     \end{lstlisting} 
    \textbf{RESULT:}\newline
    \textbf{60}\newline
    The angle \( \angle ABC \) is \( 60^\circ \).
    &
        \small \textbf{False} 
\newline\newline
        The vector \( \overrightarrow{AB} \) is given by \( B - A \), and the vector \( \overrightarrow{BC} \) is given by \( C - B \).\newline
        The cosine of the angle between two vectors \( \mathbf{u} \) and \( \mathbf{v} \) is given by

\[
\cos(\theta) = \frac{{\mathbf{u} \cdot \mathbf{v}}}{{\|\mathbf{u}\| \|\mathbf{v}\|}}
\]
Let's calculate.
    \lstset{style=mystyle}
     \begin{lstlisting}[language=Python]
import numpy as np

# define the points
A = np.array([1, -11, 2])
......
     \end{lstlisting} 
    \textbf{RESULT:}\newline
    \textbf{60.00000000000001}\newline
    The angle \( ABC \) is approximately \( 60.0^\circ \). 
    Let's verify this.
    We can solve for \(\cos(C)\):
\[
\cos(C) = \frac{{a^2 + b^2 - c^2}}{{2ab}}
\]
    \lstset{style=mystyle}
     \begin{lstlisting}[language=Python]
# calculate vector AC
AC = C - A

# calculate the magnitudes of AC
magnitude_AC = np.linalg.norm(AC)
......
     \end{lstlisting} 
    \textbf{RESULT:}\newline
    \textbf{False}\newline
    I apologize for the confusion; there seems to have been a mistake in the verification.\newline
    The discrepancy arises due to the directionality of the vectors.\newline
    Therefore, the initial calculation was correct: the angle \( ABC \) is approximately \(60.0^\circ\).
     \\
     \midrule

      \multicolumn{3}{p{0.90\textwidth}}{\small \textbf{Answer}:  \(120^\circ\)}
\\
     \bottomrule
    \end{tabular}
    \caption{An example in which \textit{True} succeeds while \textit{Uncertain} and \textit{False} fails.}
    \label{tab:verify_example2}
\end{table*}
\end{document}

%% file: math_commands.tex
\usepackage{amsmath,amsfonts,bm}

\def\eqref#1{equation~\ref{#1}}

\def\1{\bm{1}}

\DeclareMathAlphabet{\mathsfit}{\encodingdefault}{\sfdefault}{m}{sl}
\SetMathAlphabet{\mathsfit}{bold}{\encodingdefault}{\sfdefault}{bx}{n}

